\documentclass[a4paper,fleqn]{cas-sc}

\usepackage[numbers,sort&compress]{natbib}

\usepackage{bm}
\usepackage{mathtools}
\usepackage{algorithm}
\usepackage{algpseudocode}
\usepackage{tabularx}
\usepackage{xltabular}
\usepackage{textcomp}
\usepackage{pifont}
\usepackage{bbm}
\usepackage{enumitem}

\begin{document}
\let\WriteBookmarks\relax

\shorttitle{M3-Score: fidelity, memorization and coverage as separate axes}
\shortauthors{S. Nishankar et al.}

\title [mode = title]{M3-Score: Fidelity, Memorization and Coverage as Separate Axes for Evaluating Generative Radiology Image Models}

\RenewDocumentCommand \printorcid {} {}

\author[aff1]{Sathiyamohan Nishankar}

\credit{Conceptualization, Methodology, Software, Formal analysis, Investigation, Visualization, Writing -- original draft}

\author[aff3]{Pubudu Sanjeewani}
\cormark[1]
\credit{Methodology, Supervision, Validation, Writing -- review \& editing}

\author[aff4,aff5]{Asanka Perera}
\credit{Supervision, Validation, Writing -- review \& editing}

\affiliation[aff1]{organization={Faculty of Engineering, University of Peradeniya}, country={Sri Lanka}}

\affiliation[aff3]{organization={School of Computing Technologies, RMIT University}, city={Melbourne}, country={Australia}}

\affiliation[aff4]{organization={School of Engineering \& Digital Technologies, University of Southern Queensland}, city={Brisbane}, country={Australia}}

\affiliation[aff5]{organization={School of Engineering \& Technologies, UNSW}, city={Canberra}, country={Australia}}


\begin{abstract}
Quantitative evaluation of generative models for radiology remains challenging. Clinically relevant structures are often small and infrequent, feature spaces learned from natural images may represent them poorly, and a single summary score cannot distinguish limited fidelity from limited diversity. This study proposes the Medical Multi-axis Maximum Mean Discrepancy score (M3-Score), an evaluation framework based on RadioDINO-s16, a frozen vision transformer pretrained on radiology images. M3-Score reports three complementary axes computed at pre-specified encoder depths: \emph{fidelity}, measured by an unbiased multi-bandwidth radial basis function (RBF) MMD$^2$ at the final block; \emph{memorization}, measured by nearest-neighbor distances at 75\% depth; and \emph{coverage}, defined as the fraction of real images with a generated neighbor within their $k$-nearest-neighbor radius at 33\% depth. Reference sets are sampled across subjects to limit the influence of correlated slices. On BraTS brain MRI, the fidelity axis ordered five comparison sets of increasing severity (Spearman $\rho = 1.00$), and a subject-disjoint real set yielded $\mathrm{MMD}^2 = 0$ (permutation $p = 1$). An unconditional denoising diffusion probabilistic model achieved $\mathrm{MMD}^2 = 0.073$ (95\% confidence interval $[0.071, 0.080]$) but covered only 38\% of the real distribution. Under progressive mode dropping, $k$-NN manifold recall increased at all twelve encoder blocks, whereas the proposed coverage estimator decreased monotonically ($\rho = -1.00$). RadioDINO-s16 features separated real brain MRI from generated samples with a ROC-AUC of 0.819, compared with 0.555 for InceptionV3 and 0.582 for CLIP. Across a twentyfold range of sample sizes, the mean M3 value varied by a factor of 1.05, compared with 2.52 for the Fr\'{e}chet Inception Distance. Using expert glioma masks with area-matched controls, M3 was 1.4--1.8 times more responsive to tumor-region perturbations than to healthy-region perturbations, whereas an Inception-based MMD baseline yielded a ratio of approximately 1.05. M3-Score is intended for the evaluation of CT and MRI synthesis.
Code is available at \url{https://github.com/Nishan-Charlie/M3-Score/tree/main}.
\end{abstract}

\begin{keywords}
Generative models\sep Medical image synthesis\sep Radiology image generation\sep Generative model evaluation \sep Vision transformers\sep Maximum mean discrepancy
\end{keywords}
\date{}

\maketitle

\section{Introduction}
\label{sec:introduction}

Generative models are increasingly used to mitigate the scarcity of annotated medical imaging data, which remains a major constraint on the development of clinical artificial intelligence (AI)~\citep{kebaili2023deep, kazerouni2023diffusion}. The construction of large, high-quality datasets is limited by patient privacy regulations, the cost of expert annotation, class imbalance arising from rare diseases, and the practical difficulty of multi-center data sharing~\citep{finetti2025data, akpinar2024synthetic}. Generative Adversarial Networks (GANs)~\citep{goodfellow2014generative} established medical image synthesis as a practical tool, and Denoising Diffusion Probabilistic Models (DDPMs)~\citep{ho2020denoising} subsequently improved sample quality, producing realistic images in MRI, CT, chest radiography, retinal fundus photography, dermoscopy, and ultrasound~\citep{kazerouni2023diffusion, khader2023denoising, pinaya2022brain}. Synthetic data augmentation has been reported to improve segmentation, classification, and reconstruction performance in these modalities, particularly for rare pathologies~\citep{finetti2025data, akpinar2024synthetic, kebaili2023deep}.

The use of synthetic images in clinical and research pipelines requires rigorous validation. A generated medical image may obtain favorable scores on standard quantitative metrics while containing anatomically implausible structures, inconsistent intensity distributions, or hallucinated findings~\citep{kelkar2023assessing, deo2025metrics}. Such errors are apparent to domain experts but are frequently not captured by conventional evaluation measures. Evaluation metrics that are appropriate for the medical imaging domain are therefore required.

\subsection{Limitations of Existing Evaluation Metrics}
\label{sec:why_fail}

The Fr\'{e}chet Inception Distance (FID)~\citep{heusel2017gans} and the Kernel Inception Distance (KID)~\citep{binkowski2018demystifying} are the most widely used metrics for generative image models. Both compute a distance between real and generated images in the feature space of InceptionV3~\citep{szegedy2016rethinking}, a classifier pretrained on ImageNet. Table~\ref{tab:master} summarizes these and other commonly used metrics. The application of such metrics to medical images is subject to the limitations described below.

\paragraph{Domain mismatch}
InceptionV3 was trained to classify natural images of objects, animals, and scenes, and its representations are adapted to natural-image statistics. Consequently, it may be insensitive to clinically relevant properties such as tissue texture heterogeneity in MRI, Hounsfield unit distributions in CT, vascular patterns in fundus photographs, and ground-glass opacities in chest radiographs~\citep{kelkar2023assessing, skandarani2023gans}. \citet{kelkar2023assessing} trained a GAN on simulated medical images from canonical stochastic image models and showed that it attained favorable FID scores while failing to reproduce per-image statistics relevant to objective image-quality assessment. The relationship between the training domain of a backbone and the reliability of the resulting metric is, however, not straightforward. \citet{woodland2024feature} evaluated sixteen generative models across four medical imaging modalities and found that ImageNet-based feature extractors produced rankings more consistent with expert judgment than extractors trained on RadImageNet~\citep{mei2022radimagenet}, a multi-modality CT, MRI, and ultrasound dataset, whose rankings were volatile. This finding is directly relevant to the present work, because the RadioDINO encoder used here (Section~\ref{sec:proposed}) is also pretrained on RadImageNet; the RadImageNet extractors evaluated by \citet{woodland2024feature} were supervised convolutional networks, whereas RadioDINO is a self-supervised vision transformer. Domain-specific pretraining alone is therefore insufficient to establish metric reliability, and the suitability of a medical encoder must be assessed empirically.

\paragraph{Aggregation into a single score}
FID and KID summarize an entire generated set with a single scalar, which can conceal distinct failure modes. In the brain MRI experiments reported in Section~\ref{sec:results_canonical}, an unconditional DDPM is close to the real distribution in fidelity ($\mathrm{MMD}^2 = 0.073$, compared with $0$ for a held-out real set) but covers only $38\%$ of the real distribution. Its samples are individually plausible but insufficiently diverse. Because fidelity and coverage are distinct properties, a single scalar cannot characterize both.

\paragraph{Absence of per-image scores}
FID, KID, and manifold-based metrics~\citep{kynkaanniemi2019pr, naeem2020reliable} do not provide per-image scores. A single FID value does not indicate whether a generated dataset is affected by intensity drift, geometric distortion, or semantic inconsistency, nor does it identify the individual images that are suitable for downstream use.

\paragraph{Pixel-level metrics}
The Structural Similarity Index (SSIM)~\citep{wang2004image} and the peak signal-to-noise ratio (PSNR) are widely reported in medical image synthesis~\citep{dohmen2025similarity}, but both have well-documented limitations. SSIM underestimates blurring and is strongly affected by constant intensity shifts in unnormalized images~\citep{dohmen2025similarity}, and both SSIM and PSNR fail to detect local structural errors in real clinical examples across several modalities~\citep{breger2025study}. PSNR is highly sensitive to intensity normalization, and its use for evaluating synthetic MR images has been discouraged~\citep{dohmen2025similarity}. More generally, \citet{deo2025metrics} demonstrated that most standard metrics are insensitive to localized, clinically relevant alterations such as distorted tumor boundaries, and that metric-based rankings of generative models are inconsistent with downstream segmentation performance. Manifold-based metrics~\citep{kynkaanniemi2019pr, naeem2020reliable} evaluate diversity separately from fidelity, but their computational cost limits their use for monitoring during training.

\begin{table*}[t]
\centering
\caption{Commonly used evaluation metrics for generative medical image models and the proposed M3-Score. The column Dir.\ indicates whether lower ($\downarrow$) or higher ($\uparrow$) values are preferable. M3-Score reports fidelity, memorization, and coverage as separate axes, each computed at a pre-specified encoder block.}
\label{tab:master}

\renewcommand{\arraystretch}{1.15}
\setlength{\tabcolsep}{6pt}

\begin{tabularx}{\textwidth}{
    @{}
    >{\raggedright\arraybackslash}p{2.0cm}
    >{\raggedright\arraybackslash}p{3.2cm}
    >{\centering\arraybackslash}p{1.0cm}
    X
    @{}
}
\toprule
\textbf{Category} &
\textbf{Metric} &
\textbf{Dir.} &
\textbf{Measured property} \\
\midrule

\textbf{Distributional}
& FID~\citep{heusel2017gans}
& $\downarrow$
& Fr\'{e}chet distance between the feature distributions of real and generated images. \\

& KID~\citep{binkowski2018demystifying}
& $\downarrow$
& Unbiased kernel-based distance between distributions, with improved small-sample behavior. \\

& IS~\citep{salimans2016is}
& $\uparrow$
& Classifier-based measure of sample quality and diversity. \\

\midrule

\textbf{Perceptual}
& SSIM~\citep{wang2004image}
& $\uparrow$
& Structural similarity based on luminance, contrast, and structure. \\

& PSNR
& $\uparrow$
& Pixel-level reconstruction quality expressed as peak signal-to-noise ratio. \\

& MS-SSIM
& $\uparrow$
& Structural similarity computed at multiple scales. \\

& LPIPS~\citep{zhang2018lpips}
& $\downarrow$
& Perceptual distance in the feature space of a pretrained deep network. \\

\midrule

\textbf{Manifold}
& $\alpha$-Precision~\citep{kynkaanniemi2019pr,alaa2022faithful}
& $\uparrow$
& Fidelity of generated samples with respect to the real-data manifold. \\

& $\beta$-Recall~\citep{kynkaanniemi2019pr,alaa2022faithful}
& $\uparrow$
& Extent to which the generated distribution covers the real-data manifold. \\

& Authenticity~\citep{alaa2022faithful}
& $\uparrow$
& Degree to which generated samples differ from training examples. \\

& Coverage~\citep{naeem2020reliable}
& $\uparrow$
& Fraction of real samples with a nearby generated sample. \\

& Density~\citep{naeem2020reliable}
& $\uparrow$
& Concentration of generated samples around the real-data manifold. \\

\midrule

\textbf{Proposed}
& \textbf{M3 Fidelity}
& $\downarrow$
& Multi-bandwidth RBF MMD$^2$ between real and generated representations. \\

& \textbf{M3 Memorization}
& $\downarrow$
& Fraction of generated samples that lie unusually close to a real sample. \\

& \textbf{M3 Coverage}
& $\uparrow$
& Fraction of real samples covered by the generated distribution. \\

\bottomrule
\end{tabularx}
\end{table*}

\subsection{Proposed Framework}
\label{sec:proposed}

This study proposes the \textbf{Medical Multi-axis Maximum Mean Discrepancy score (M3-Score)}, an evaluation framework for generative radiology image models. The framework addresses the limitations described above through three design elements.

\paragraph{Radiology-pretrained representation}
M3-Score uses frozen Vision Transformer (ViT) features from RadioDINO~\citep{zedda2025radiodino}, a self-supervised encoder pretrained on RadImageNet~\citep{mei2022radimagenet}, which comprises approximately 1.35 million CT, MRI, and ultrasound images. The compact \textbf{RadioDINO-s16} variant (ViT-S/16, 12 transformer blocks, 384-dimensional embeddings) is used throughout. When the scoring rule is held fixed and only the feature space is varied (Section~\ref{sec:results_ood}), RadioDINO-s16 separates real brain MRI from DDPM samples with a ROC-AUC of $0.819$, whereas the InceptionV3 features underlying FID ($0.555$) and the CLIP features underlying CMMD ($0.582$) perform close to chance.

\paragraph{Separate evaluation axes}
Different generation failures are expressed at different representational depths. M3-Score therefore reports three separately defined axes, each computed at an encoder block pre-specified according to its relative depth. The \textbf{fidelity} axis is an unbiased multi-bandwidth radial basis function (RBF) MMD$^2$~\citep{gretton2012kernel} computed at the final block (L12), analogous to the use of the penultimate InceptionV3 layer in FID. The \textbf{memorization} axis is based on nearest-neighbor distances at $75\%$ depth (L9), where instance-level variation is retained. The \textbf{coverage} axis is computed at $33\%$ depth (L4), where spatial and structural variation is encoded, as the fraction of real images with a generated neighbor inside their $k$-nearest-neighbor ($k$-NN) radius~\citep{naeem2020reliable}. The neighborhood radii are estimated on the real set. As shown in Section~\ref{sec:results_coverage}, radii estimated on the generated set, as in $k$-NN manifold recall~\citep{kynkaanniemi2019pr}, cause the statistic to increase under mode dropping.

\paragraph{Statistical reporting}
The fidelity axis is reported with a permutation-test $p$-value and a bootstrap 95\% confidence interval (CI), which FID, KID, and CMMD do not provide by default. For $N = 500$ images, the complete evaluation requires approximately $8$\,s on a single NVIDIA A6000 GPU (Section~\ref{sec:method_complexity}).

\subsection{Contributions}
\label{sec:contributions}

The main contributions of this study are as follows.

\begin{enumerate}[leftmargin=*,label=\arabic*.,itemsep=0.3em]

  \item \textbf{A three-axis evaluation framework.}
        M3-Score reports fidelity, memorization, and coverage as separate quantities computed from RadioDINO-s16 features at pre-specified encoder depths, with a permutation test and bootstrap CI for the fidelity axis.

  \item \textbf{A coverage estimator with the correct response to mode dropping.}
        Coverage based on real-set neighborhood radii decreases monotonically under mode dropping at all twelve encoder blocks ($\rho = -1.00$). Under the same perturbation, $k$-NN manifold recall increases at every block, because its radii are estimated on the subsampled generated set.

  \item \textbf{A subject-level protocol for reference-set construction.}
        Reference slices are sampled round-robin across subjects. Under this protocol, a subject-disjoint real set yields no detectable distributional shift.

  \item \textbf{An empirical comparison of feature spaces.}
        Radiology-pretrained features separate real and generated brain MRI more effectively than InceptionV3 and CLIP features. In addition, CMMD reverses the ordering of a cross-modality CT set and a weaker same-modality generator, whereas M3 and FID preserve the expected ordering.

  \item \textbf{Perturbation-based validation of each axis.}
        Each axis is evaluated against a perturbation that it is designed to detect, including an area-matched lesion-specificity analysis with expert glioma masks and a kernel-matched baseline that isolates the effect of the backbone.

  \item \textbf{An analysis of sample-size dependence.}
        Across a twentyfold range of sample sizes, the mean M3 value varies by a factor of $1.05$, whereas FID varies by a factor of $2.52$.

\end{enumerate}

\subsection{Scope}
\label{sec:scope}

The primary validation domain is axial brain MRI from BraTS 2021~\citep{menze2014multimodal,bakas2018identifying,baid2021rsna}, selected for its status as an established benchmark, its size ($38{,}781$ slices from $1{,}251$ subjects), and the diversity of tumor appearance that it contains. Additional experiments use lower-grade glioma MRI with expert annotations~\citep{buda2019association}, synthetic chest CT from a generative model trained on LIDC-IDRI~\citep{armato2011lidc}, and synthetic retinal fundus images as a non-radiology domain.

M3-Score is intended for radiology modalities represented in the pretraining data of RadioDINO, in particular CT and MRI. For images outside this domain, M3 values reflect distances in a radiology-specific representation and should not be interpreted as general measures of visual distance (Section~\ref{sec:results_ood}); applications outside radiology would require a different backbone. M3-Score is intended to complement, rather than replace, established metrics such as FID~\citep{heusel2017gans} and SSIM~\citep{wang2004image}.

\subsection{Organization}
\label{sec:organization}

Section~\ref{sec:related} reviews related work. Section~\ref{sec:methodology} describes the proposed framework. Section~\ref{sec:results} presents the experimental setup and results. Section~\ref{sec:discussion} discusses the findings and limitations, and Section~\ref{sec:conclusion} concludes the paper. Appendix~\ref{app:cka} specifies the layer-redundancy analysis.

\section{Related Work}
\label{sec:related}

Evaluation metrics for generative image models can be grouped into pixel-level and perceptual metrics, distributional metrics based on pretrained networks, manifold-based metrics, and metrics developed specifically for medical images. Each group is reviewed below, followed by an overview of self-supervised radiology encoders.

\subsection{Pixel-Level and Perceptual Metrics}
\label{sec:related_pixel}

The earliest image quality metrics operate directly on pixel intensities. PSNR is defined as the logarithm of the ratio between the maximum possible signal power and the mean squared error, which makes it sensitive to normalization and inconsistent across intensity ranges. \citet{wang2004image} introduced SSIM, which compares luminance, contrast, and structure within local windows. SSIM became the standard paired-image metric in medical image synthesis~\citep{dohmen2025similarity} and remains widely reported alongside distributional metrics. However, \citet{breger2025study} documented real-world MRI, CT, X-ray, and other medical imaging cases in which PSNR and SSIM misjudge image quality, including failures to detect local structural errors, and \citet{dohmen2025similarity} showed that both metrics underestimate blurring and that PSNR depends strongly on the intensity normalization applied. \citet{zhang2018lpips} proposed the Learned Perceptual Image Patch Similarity (LPIPS), which measures distance in the feature space of a pretrained deep network. LPIPS agrees more closely with human judgment than PSNR or SSIM on natural images, but it inherits the ImageNet bias of its backbone and requires paired images, which limits its applicability to the evaluation of unpaired generated sets.

\subsection{Distributional Metrics}
\label{sec:related_distributional}

The Inception Score (IS) \citep{salimans2016is} was the first widely adopted metric for evaluating unpaired generative models. It applies a pretrained InceptionV3 classifier \citep{szegedy2016rethinking} to generated images and computes the Kullback-Leibler divergence between the conditional and marginal label distributions. Because IS is computed exclusively on generated images without comparison to a real reference set, it struggles to reliably detect mode collapse or distributional shifts.

FID \citep{heusel2017gans} addressed this limitation by comparing real and generated datasets directly. It fits multivariate Gaussian distributions to the InceptionV3 features of both sets and computes the Fr\'{e}chet distance between them. FID quickly became the standard benchmark for generative models and remains the most frequently reported metric in medical image synthesis. However, it presents three critical limitations for medical applications: it assumes features follow a Gaussian distribution (an assumption regularly violated by ReLU activations), it is a biased estimator heavily dependent on sample size, and its backbone encodes natural image statistics rather than relevant clinical features \citep{kelkar2023assessing, skandarani2023gans}.

KID \citep{binkowski2018demystifying} mitigated the statistical flaws of FID by replacing the Fr\'{e}chet distance with an unbiased estimator of the Maximum Mean Discrepancy (MMD) using a polynomial kernel. Formalized by \citet{gretton2012kernel} as a nonparametric two-sample test statistic, MMD measures the maximum difference in expectations between two distributions within a reproducing kernel Hilbert space. Because KID avoids parametric assumptions and is unbiased, it is far more reliable than FID at the small sample sizes typical of medical imaging. Nevertheless, it still relies on the InceptionV3 backbone, meaning it inherits the same domain mismatch as FID.

More recently, \citet{jayasumana2024cmmd} proposed CLIP Maximum Mean Discrepancy (CMMD) as a modern alternative. CMMD computes an unbiased Gaussian RBF MMD using CLIP embeddings \citep{radford2021clip}. Because CLIP is trained on a massive and diverse dataset of paired images and text captions, its embeddings capture richer semantic content than InceptionV3. The authors demonstrated that while FID can contradict human evaluators, fail to reflect gradual quality improvements, and even artificially improve under certain image distortions, CMMD behaves consistently. Although CMMD is distribution-free, unbiased, and sample-efficient, CLIP is fundamentally trained on natural images. Consequently, its embeddings lack the domain-specific knowledge required to accurately evaluate the intricate anatomical and pathological structures present in medical imaging.

\subsection{Manifold-Based Metrics}
\label{sec:related_manifold}

A complementary line of work seeks to disentangle the fidelity and diversity of generated images, two critical aspects that FID convolutes into a single score. \citet{kynkaanniemi2019pr} introduced improved precision and recall metrics by approximating the real and generated data manifolds using $k$-nearest-neighbor hyperspheres. In this framework, precision represents the fraction of generated samples that fall within the real manifold, while recall measures the fraction of real samples covered by the generated manifold.

However, \citet{naeem2020reliable} demonstrated that these initial metrics are highly sensitive to outliers and fail to reach their maximum values even when comparing identical distributions. To resolve this, they proposed density and coverage as more robust alternatives. Expanding on this diagnostic approach, \citet{alaa2022faithful} introduced $\alpha$-precision, $\beta$-recall, and authenticity, with the latter specifically designed to quantify how closely generated samples replicate the original training data. While manifold-based metrics offer highly valuable diagnostic insights, constructing a reliable manifold requires dense sampling, which significantly increases the computational cost of the evaluation.

\subsection{Metrics for Medical Images}
\label{sec:related_medical}

The well-documented limitations of ImageNet-pretrained features in the medical domain have spurred the development of specialized evaluation metrics. As established in Section~\ref{sec:why_fail}, standard metrics often overlook clinically critical errors \citep{kelkar2023assessing, deo2025metrics}. The most relevant prior approach to our proposed M3-Score is the Fr\'{e}chet Radiomic Distance (FRD) introduced by \citet{konz2025frd}. To bridge the domain gap, FRD abandons deep features in favor of standardized radiomic descriptors, computing the Fr\'{e}chet distance between Gaussian distributions fitted to the real and generated sets. This approach successfully outperformed FID, KID, and CMMD across out-of-distribution (OOD) detection, image-to-image translation, and unconditional generation. Furthermore, it demonstrated stability at small sample sizes and correlated strongly with radiologist quality assessments.

While M3-Score shares the goal of domain-specific evaluation, it diverges from FRD in three fundamental ways. First, by utilizing deep features from a radiology-pretrained ViT rather than hand-crafted radiomics, M3-Score captures higher-order anatomical structures that first-order and texture-based descriptors intrinsically miss. Second, rather than collapsing the evaluation into a single scalar, M3-Score isolates three distinct generative axes at predetermined representational layers: fidelity (L12), memorization (L9), and coverage (L4). Finally, the fidelity axis of M3-Score calculates an unbiased multi-bandwidth RBF $MMD^2$ equipped with a permutation $p$-value and bootstrap confidence intervals, providing the robust statistical significance testing absent from FRD's standard Fr\'{e}chet distance formulation.


\subsection{Self-Supervised Radiology Encoders}
\label{sec:related_encoders}

RadioDINO \citep{zedda2025radiodino} comprises a family of self-supervised Vision Transformer (ViT) encoders pretrained on RadImageNet \citep{mei2022radimagenet}, a comprehensive dataset containing approximately 1.35 million CT, MRI, and ultrasound images across 11 anatomical regions. Following the DINO/DINOv2 self-supervised learning paradigm \citep{oquab2023dinov2}, RadioDINO requires no text supervision. Its learned representations have demonstrated strong capabilities across downstream classification, segmentation, and interpretability tasks \citep{zedda2025radiodino}. Similar domain-specific efforts include RAD-DINO \citep{perezrad}, which was trained exclusively on chest radiographs. 

Crucially, \citet{woodland2024feature} demonstrated that the choice of feature extractor fundamentally dictates the alignment between generative evaluation metrics and expert clinical judgment. Furthermore, they cautioned that medical pretraining alone does not automatically guarantee closer agreement with human raters: in their study, Fr\'{e}chet distances computed with RadImageNet-trained extractors, the same pretraining corpus as RadioDINO, produced model rankings that were volatile and inconsistent with expert judgment. Motivated by these findings, we directly compare the RadioDINO-s16 representation against established InceptionV3 and CLIP baselines in Section~\ref{sec:results_ood}.

In summary, the existing literature reveals three critical gaps in the evaluation of medical generative models: (a) a persistent reliance on general-purpose encoders that fail to capture radiology-specific structures; (b) the use of single scalar scores that conflate distinct failure modes, such as low fidelity, memorization, and restricted coverage; and (c) a distinct lack of rigorous statistical significance testing and per-image scoring mechanisms. M3-Score systematically addresses these deficiencies by leveraging radiology-pretrained features to construct three independent evaluation axes (fidelity, memorization, and coverage) computed at predetermined network depths, alongside robust statistical reporting for the fidelity axis.

\section{Methodology}
\label{sec:methodology}

\begin{figure}[pos=htbp]
\centering
\includegraphics[width=\textwidth]{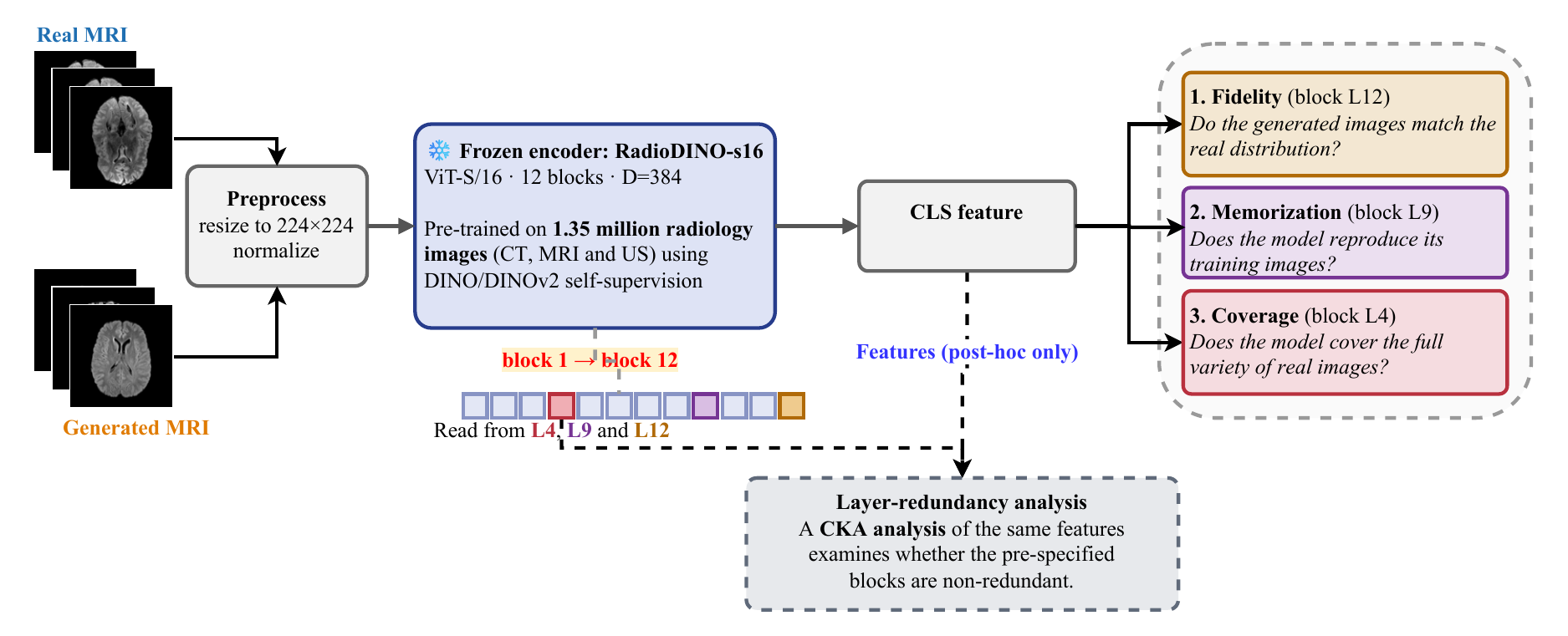}
\caption{Overview of the $M^3$-Score framework. Real and generated images are preprocessed and encoded by a frozen RadioDINO-s16 encoder. The [CLS] token features from three pre-specified blocks are used to compute fidelity (L12), memorization (L9), and coverage (L4). The dashed path denotes the layer-redundancy analysis based on centered kernel alignment (CKA), which is used solely for analysis and does not contribute to the reported axes.}
\label{fig:methodology_pipeline}
\end{figure}

This section describes the construction of reference sets, feature extraction, the three evaluation axes, the choice of encoder depths, the per-image score, and the computational cost of the framework. Figure~\ref{fig:methodology_pipeline} and Algorithm~\ref{alg:m3canonical} summarize the procedure.

\subsection{Notation and Overview}
\label{sec:method_overview}

Let $\mathcal{X}_r = \{x_i\}_{i=1}^{N_r}$ denote a reference set of $N_r$ real radiology images and $\mathcal{X}_g = \{\hat{x}_j\}_{j=1}^{N_g}$ a set of $N_g$ images produced by the generative model under evaluation. Let $\Phi$ denote the frozen RadioDINO-s16 encoder comprising $L = 12$ transformer blocks, and let $E_r^{(\ell)} \in \mathbb{R}^{N_r \times D}$ and $E_g^{(\ell)} \in \mathbb{R}^{N_g \times D}$ denote the matrices of [CLS] token embeddings for the real and generated sets at block $\ell$, with a hidden dimension of $D = 384$. $M^3$-Score reports three evaluation axes computed at pre-specified blocks: fidelity at $\ell_{\text{fid}} = 12$, memorization at $\ell_{\text{mem}} = 9$, and coverage at $\ell_{\text{cov}} = 4$. The complete computational procedure is detailed in Algorithm~\ref{alg:m3canonical}.

\begin{algorithm}[t]
\caption{$M^3$-Score}
\label{alg:m3canonical}
\begin{algorithmic}[1]
\Require Real set $\mathcal{X}_r$ of size $N_r$ (sampled across subjects, \S\ref{sec:method_reference}),
         generated set $\mathcal{X}_g$ of size $N_g$, frozen encoder $\Phi$, pre-specified depths
         $\ell_{\text{fid}}{=}12,\ \ell_{\text{mem}}{=}9,\ \ell_{\text{cov}}{=}4$
\Ensure  Fidelity $(\widehat{\text{MMD}}^2, p, [\text{CI}])$, memorization rate, coverage
\For{$\ell \in \{12,\,9,\,4\}$}
  \State $E_r^{(\ell)} \leftarrow \textsc{CLS}(\Phi,\mathcal{X}_r,\ell)$
  \State $E_g^{(\ell)} \leftarrow \textsc{CLS}(\Phi,\mathcal{X}_g,\ell)$
         \Comment{[CLS] tokens; \S\ref{sec:method_backbone}}
\EndFor
\State \textit{// Fidelity axis (L12)}
\State $\widehat{\text{MMD}}^2 \leftarrow \textsc{MMD2U}(E_r^{(12)}, E_g^{(12)})$
       \Comment{Alg.~\ref{alg:mmd}}
\State $p \leftarrow \textsc{PermTest}(E_r^{(12)}, E_g^{(12)};\,B{=}500)$
\State $[\text{CI}] \leftarrow \textsc{Bootstrap}(E_r^{(12)}, E_g^{(12)};\,B{=}500)$
\State \textit{// Memorization axis (L9)}
\State $\theta \leftarrow$ median within-real $1$-NN distance
\State $\text{mem} \leftarrow \frac{1}{N_g}\sum_{j=1}^{N_g}\mathbb{1}\!\left[\min_i \lVert E_{g,j}^{(9)} - E_{r,i}^{(9)}\rVert_2 < \theta\right]$
\State \textit{// Coverage axis (L4), radii estimated on the real set}
\State $r_i \leftarrow \textsc{NND}_k(E_{r,i}^{(4)}, E_r^{(4)})$ for each $i$
       \Comment{$k{=}5$; excluding self}
\State $\text{cov} \leftarrow \frac{1}{N_r}\sum_{i=1}^{N_r}\mathbb{1}\!\left[\min_j \lVert E_{r,i}^{(4)} - E_{g,j}^{(4)}\rVert_2 \le r_i\right]$
\State \Return $\{\widehat{\text{MMD}}^2, p, [\text{CI}]\},\ \text{mem},\ \text{cov}$
\end{algorithmic}
\end{algorithm}

\subsection{Reference Set Construction}
\label{sec:method_reference}

The effectiveness of any distributional metric relies as much on the quality of the reference set as it does on the generative model. In volumetric medical imaging, adjacent axial slices from a single patient scan are highly correlated. Consequently, a reference set of $N$ slices does not represent $N$ independent observations. Instead, its true information content is dictated by the number of unique subjects included. Naively selecting the first $N$ files in sorted order is especially problematic because public datasets are typically structured by subject identifiers. This flawed approach captures very few distinct patients and, in multi-institutional datasets like BraTS, restricts the data to a limited number of acquisition sites, scanner models, and imaging protocols.

To resolve this, all real reference sets in this study are constructed using a round-robin sampling strategy across subjects. Patients are first randomly ordered, and exactly one slice is drawn from each subject before a second slice is taken from anyone. Furthermore, when comparing two real datasets, they are strictly drawn from disjoint, non-overlapping groups of subjects. Our standard reference set consists of $N = 500$ slices derived from 500 distinct individuals.

This sampling protocol aligns with established best practices in discriminative medical image analysis, where data must be partitioned at the subject level to prevent highly correlated slices from leaking across training and test sets \citep{varoquaux2022machine}. It also mitigates the well-documented issue of scanner-specific signatures dominating multi-site datasets \citep{glocker2019machine}. Because the reference set in generative evaluation serves an analogous role to the test set in discriminative tasks, maintaining this rigorous subject-level separation is essential. For maximum transparency, we explicitly report the number of distinct subjects alongside the total slice count $N$ for every evaluated dataset.

\subsection{Feature Extraction}
\label{sec:method_backbone}

RadioDINO \citep{zedda2025radiodino} is a family of Vision Transformer (ViT) encoders \citep{dosovitskiy2021an} pretrained on the RadImageNet dataset. These models are trained using DINO and DINOv2 self-supervised learning frameworks \citep{oquab2023dinov2}. In this work, we utilize the RadioDINO-s16 architecture. This specific model consists of $L = 12$ transformer blocks, a hidden dimension of $D = 384$, and six attention heads.

During preprocessing, each input image is resized to $224 \times 224$ pixels. The image is then divided into 196 non-overlapping $16 \times 16$ patches. A [CLS] token is prepended to these patches, resulting in a final token sequence length of 197. All encoder weights remain strictly frozen during feature extraction.

Mathematically, for a given input image $x$, let $\mathbf{H}^{(\ell)}(x) \in \mathbb{R}^{197 \times D}$ represent the hidden states at transformer block $\ell$. We extract the feature of $x$ at block $\ell$ using the [CLS] token: $f^{(\ell)}(x) = \mathbf{H}^{(\ell)}_{0}(x) \in \mathbb{R}^{D}$. This serves as the standard image-level representation for the ViT. By stacking the [CLS] features of all $N$ images within a dataset, we obtain the complete embedding matrix $E^{(\ell)} \in \mathbb{R}^{N \times D}$.

\paragraph{Feature Normalization}
The hidden states within RadioDINO blocks exhibit exceptionally large $\ell_2$ norms, typically ranging between $10^2$ and $3 \times 10^2$ depending on the network depth. If left uncorrected, these large magnitudes dominate the pairwise distances computed within the Gaussian kernel. This distortion skews the median-heuristic bandwidth and ultimately makes $MMD^2$ values incomparable across different layers or datasets.

To resolve this scale dependency, every embedding is projected onto the unit sphere prior to kernel evaluation:
\begin{equation}
  \label{eq:l2norm}
  \hat{e} = \frac{e}{\lVert e \rVert_2 + \varepsilon}, \quad \text{where } \varepsilon = 10^{-8}.
\end{equation}
Following this projection, the norm becomes strictly $\lVert \hat{e} \rVert_2 = 1$. Consequently, the inner product $\hat{u}^\top \hat{v}$ naturally equates to the cosine similarity between $u$ and $v$, safely bounding the values within the $[-1, 1]$ interval.

\subsection{Fidelity Axis}
\label{sec:method_mmd}

\paragraph{Kernel}
Let $\hat{R} \in \mathbb{R}^{N_r \times D}$ and $\hat{G} \in \mathbb{R}^{N_g \times D}$ denote the $\ell_2$-normalized real and generated embedding matrices at layer L12. The fidelity axis measures the discrepancy between $\hat{R}$ and $\hat{G}$ using the Maximum Mean Discrepancy (MMD) \citep{gretton2012kernel}. We compute this under a multi-bandwidth Gaussian RBF kernel. For two unit vectors $u, v \in \mathbb{R}^D$, the kernel is defined as:
\begin{equation}
  \label{eq:kernel}
  k(u, v) = \frac{1}{|\mathcal{B}|}
            \sum_{\sigma^2 \in \mathcal{B}}
            \exp\!\Bigl(
              -\frac{\lVert u - v \rVert_2^2}{2\sigma^2}
            \Bigr).
\end{equation}
This formulation represents the average of five distinct Gaussian kernels. Their squared bandwidths are given by:
\begin{equation}
  \label{eq:bandwidths}
  \mathcal{B} = \bigl\{\tfrac14,\ \tfrac12,\ 1,\ 2,\ 4\bigr\}
                \cdot \sigma_\mathrm{med}^2 .
\end{equation}
We determine the base value $\sigma_\mathrm{med}^2$ using the median heuristic \citep{gretton2012kernel}. It is calculated as the median squared Euclidean distance across all distinct pairs in the pooled sample $\hat{R} \cup \hat{G}$. Averaging over a geometric sequence of bandwidths ensures the kernel captures differences at multiple scales. It also eliminates the need to manually select a single bandwidth. Concurrently, the median heuristic automatically scales the kernel to match the data distribution. Because the embeddings lie on the unit sphere, the squared distance simplifies to $\lVert u - v \rVert_2^2 = 2(1 - u^\top v) \in [0,4]$. Consequently, every kernel evaluation falls strictly within $(0, 1]$, ensuring the $MMD^2$ remains bounded, similar to KID \citep{binkowski2018demystifying}.

\paragraph{Unbiased Estimator}
We define the kernel matrices as $K^{rr}_{ij} = k(\hat{R}_{i,:}, \hat{R}_{j,:})$, $K^{gg}_{ij} = k(\hat{G}_{i,:}, \hat{G}_{j,:})$, and $K^{rg}_{ij} = k(\hat{R}_{i,:}, \hat{G}_{j,:})$. The unbiased estimator for the squared MMD is then:
\begin{equation}
  \label{eq:mmd}
  \widehat{\mathrm{MMD}}^2 =
  \frac{1}{N_r(N_r-1)} \sum_{i \neq j} K^{rr}_{ij}
  + \frac{1}{N_g(N_g-1)} \sum_{i \neq j} K^{gg}_{ij}
  - \frac{2}{N_r N_g} \sum_{i,j} K^{rg}_{ij}.
\end{equation}
Due to linearity, this estimator perfectly equals the average of the individual single-bandwidth estimators. Finite-sample variation can occasionally produce negative values. We set these to zero. This clipping introduces only a negligible positive bias when the true population $MMD^2$ is near zero. Prior to clipping, Equation~\eqref{eq:mmd} serves as a strictly unbiased estimator of the population $MMD^2$. Because the Gaussian kernel is characteristic \citep{gretton2012kernel}, this population value is exactly zero if and only if the two distributions are identical. Unlike FID, this approach makes no parametric assumptions about how the embeddings are distributed. The exact computational steps are detailed in Algorithm~\ref{alg:mmd}.

\begin{algorithm}[t]
\caption{Unbiased multi-bandwidth RBF $MMD^2$ ($\textsc{MMD2U}$)}
\label{alg:mmd}
\begin{algorithmic}[1]
\Require Real and generated embedding matrices at a single layer,
         $E_r \in \mathbb{R}^{N_r \times D}$, $E_g \in \mathbb{R}^{N_g \times D}$;
         bandwidth scales $\mathcal{S} = \{0.25, 0.5, 1, 2, 4\}$
\Ensure  $\widehat{\mathrm{MMD}}^2 \in [0, \infty)$
\State $\hat{E}_{r,i,:} \leftarrow E_{r,i,:} / (\lVert E_{r,i,:} \rVert_2 + \varepsilon)$
\State $\hat{E}_{g,j,:} \leftarrow E_{g,j,:} / (\lVert E_{g,j,:} \rVert_2 + \varepsilon)$
       \Comment{Eq.~\eqref{eq:l2norm}}
\State $D^{rr}_{ij} \leftarrow \lVert \hat{E}_{r,i} - \hat{E}_{r,j} \rVert_2^2$
\State $D^{gg}_{ij} \leftarrow \lVert \hat{E}_{g,i} - \hat{E}_{g,j} \rVert_2^2$
\State $D^{rg}_{ij} \leftarrow \lVert \hat{E}_{r,i} - \hat{E}_{g,j} \rVert_2^2$
\State $\sigma_\mathrm{med}^2 \leftarrow
       \operatorname{median}\bigl(\{D^{\cup}_{ij} : i < j\}\bigr)$
       \Comment{pooled sample, Eq.~\eqref{eq:bandwidths}}
\State $A \leftarrow 0$
\For{each scale $s \in \mathcal{S}$}
  \State $\gamma \leftarrow 1 / (2\, s\, \sigma_\mathrm{med}^2)$
  \State $K^{rr} \leftarrow \exp(-\gamma D^{rr})$
  \State $K^{gg} \leftarrow \exp(-\gamma D^{gg})$
  \State $K^{rg} \leftarrow \exp(-\gamma D^{rg})$
  \State $T_{rr} \leftarrow
         \bigl(\sum_{i,j} K^{rr}_{ij} - \sum_i K^{rr}_{ii}\bigr) / (N_r(N_r-1))$
         \Comment{diagonal excluded}
  \State $T_{gg} \leftarrow
         \bigl(\sum_{i,j} K^{gg}_{ij} - \sum_i K^{gg}_{ii}\bigr) / (N_g(N_g-1))$
  \State $T_{rg} \leftarrow \bigl(\sum_{i,j} K^{rg}_{ij}\bigr) / (N_r N_g)$
  \State $A \leftarrow A + (T_{rr} + T_{gg} - 2\,T_{rg})$
\EndFor
\State $\widehat{\mathrm{MMD}}^2 \leftarrow
       \max(A / |\mathcal{S}|,\; 0)$
       \Comment{Eq.~\eqref{eq:mmd}}
\State \Return $\widehat{\mathrm{MMD}}^2$
\end{algorithmic}
\end{algorithm}

\paragraph{Statistical Inference}
Three statistical quantities accompany the fidelity estimate. First, we compute a permutation test $p$-value. We achieve this by pooling $E_r^{(12)}$ and $E_g^{(12)}$, randomly reassigning the dataset labels $B = 500$ times, and recalculating the estimator for each permutation. This creates a one-sided test for the null hypothesis $H_0: \mathrm{MMD}^2 = 0$. We calculate the final $p$-value using the estimator $\hat{p} = (b + 1)/(B + 1)$, where $b$ represents the number of permuted statistics that are greater than or equal to the observed value. Consequently, the smallest attainable $p$-value is $1/501 \approx 0.002$.

Second, we generate a bootstrap 95\% confidence interval (CI) from 500 independent resamples of the real and generated sets with replacement. Because each patient contributes exactly one slice to a real set, resampling real images perfectly equates to resampling patients. The population $MMD^2$ is strictly non-negative, and our estimate is clipped at zero. Therefore, this CI describes the precision of the estimate rather than acting as a formal hypothesis test against zero.

Third, we calculate the null-standardized separation $Z = (\widehat{\mathrm{MMD}}^2 - \mu_0)/\sigma_0$. Here, $\mu_0$ and $\sigma_0$ represent the mean and standard deviation of the permutation distribution. $Z$ functions as a descriptive quantity, not a standard Gaussian $z$-score. When the two dataset distributions fundamentally differ, the observed statistic converges to a positive constant. Meanwhile, $\sigma_0$ decreases at a rate of $O(1/N)$ based on sample size. As a result, $Z$ increases approximately linearly as sample size grows. Furthermore, the permutation distribution itself is highly non-Gaussian. For these critical reasons, $Z$ values should never be directly compared across different sample sizes.

\subsection{Memorization Axis}
\label{sec:method_memorization}

The memorization axis identifies generated images that are unusually close to the training data. We compute this using L9 features. For each generated image, we calculate its minimum Euclidean distance to the nearest real image: $d_j = \min_i \lVert E_{g,j}^{(9)} - E_{r,i}^{(9)} \rVert_2$. Next, we define a threshold $\theta$ as the median distance between a real image and its nearest real neighbor within the reference set. 

The memorization rate is then calculated as:
\begin{equation}
\label{eq:memorization}
\mathrm{Mem} = \frac{1}{N_g}\sum_{j=1}^{N_g} \mathbb{1}\!\left[d_j < \theta\right].
\end{equation}
This formulation measures the fraction of generated images that fall closer to a real image than a typical real image falls to its nearest neighbor. Consequently, generated images meeting this criterion are highly likely to be near-duplicates of the real data.

\subsection{Coverage Axis}
\label{sec:method_coverage}

We define coverage using L4 features. It represents the fraction of real images that contain at least one generated image within their $k$-nearest-neighbor ($k$-NN) radius. Following \citet{naeem2020reliable}, we set $k = 5$ and estimate this radius strictly within the real dataset:
\begin{equation}
\label{eq:coverage}
\mathrm{Cov} = \frac{1}{N_r}\sum_{i=1}^{N_r}
  \mathbb{1}\!\left[\;\min_j \lVert E_{r,i}^{(4)} - E_{g,j}^{(4)} \rVert_2
  \;\le\; \mathrm{NND}_k\!\left(E_{r,i}^{(4)}\right)\right],
\end{equation}
where $\mathrm{NND}_k(E_{r,i}^{(4)})$ denotes the distance to the $k$-th nearest neighbor of image $i$ within the real set. A low coverage score directly indicates that the generator fails to populate certain regions of the real data distribution. This phenomenon is commonly referred to as mode dropping.

Our choice of this specific estimator is deliberately motivated by how neighborhood radii behave under mode dropping. The traditional $k$-NN manifold recall \citep{kynkaanniemi2019pr} centers its hyperspheres on the generated samples. It sets their radii equal to the $k$-NN distances within the generated set itself. As a result, when a model drops modes and generates less diverse samples, the remaining generated samples become more widely separated. Their internal $k$-NN distances subsequently increase, which artificially enlarges the hyperspheres. Paradoxically, this means traditional recall can actually increase as generated diversity decreases.

Equation~\eqref{eq:coverage} completely avoids this flaw by centering the hyperspheres on the real samples. Because these radii depend exclusively on the real reference set, they remain completely unaffected by the generator's mode dropping. Removing generated samples can only leave real samples uncovered. Therefore, our coverage estimator remains strictly non-increasing under mode dropping. We empirically evaluate and validate both estimators in Section~\ref{sec:results_coverage}.

\subsection{Choice of Encoder Depths}
\label{sec:method_axes}

We establish the three extraction depths as fixed configuration constants based on their relative positions within the encoder architecture:
\begin{equation}
\ell_{\text{fid}} = 12, \qquad
\ell_{\text{mem}} = 9, \qquad
\ell_{\text{cov}} = 4 .
\end{equation}

Layer L12 represents the deepest and most semantically rich block. This makes it the conventional choice for computing distributional distances, directly corresponding to the use of the penultimate InceptionV3 layer in standard FID. Layer L9, located at $75\%$ network depth, preserves the instance-level variation necessary to reliably detect near-duplicates for the memorization axis. Layer L4, located at $33\%$ network depth, retains the fundamental spatial and structural variations critical for assessing generation diversity and coverage.

By strictly pre-specifying these depths, we ensure that no layer, threshold, or kernel bandwidth is retroactively selected based on favorable evaluation outcomes. Importantly, we did not deliberately optimize these depths for any specific downstream task, which prevents the metric from overfitting to a particular dataset. We empirically examine the sensitivity of these depth choices in Section~\ref{sec:results_depth}.

\subsection{Per-Image Score}
\label{sec:method_perim}

The fidelity axis summarizes an entire generated dataset. However, a per-image score allows individual generated images to be ranked based on their specific distance from the true data distribution. Standard metrics like FID, KID, and FRD inherently lack this granular capability. To maintain consistency with the fidelity axis, we compute this per-image score using the extracted L12 features. 

For a specific generated image $\hat{x}_j$ with a normalized feature vector $\hat{E}_{g,j}^{(12)}$, we define the score as its Euclidean distance to the centroid of the normalized real features:
\begin{equation}
\label{eq:perim}
s(\hat{x}_j)
=
\left\|
\hat{E}_{g,j}^{(12)} - \hat{\mu}_{r}^{(12)}
\right\|_2,
\qquad
\hat{\mu}_{r}^{(12)}
=
\frac{1}{N_r}
\sum_{i=1}^{N_r}
\hat{E}_{r,i}^{(12)}.
\end{equation}
A large value of $s(\hat{x}_j)$ directly indicates that the image lies far from the real distribution within the semantic feature space of L12. Because this calculation reuses the features already extracted for the fidelity axis, it introduces negligible computational overhead. Furthermore, we leverage this exact score for the per-image discrimination tasks evaluated in Section~\ref{sec:results_ood}.

\subsection{Computational Cost and Settings}
\label{sec:method_complexity}

Feature extraction inherently dominates the total computational cost. It requires one forward pass per image, yielding a self-attention complexity of $O(197^2 \cdot D)$ per block. The $MMD^2$ calculation requires $O(N^2)$ kernel evaluations. Similarly, the nearest-neighbor computations for the memorization and coverage axes scale at $O(N^2)$. At a standard sample size of $N = 500$, these metric calculations remain extremely fast relative to the initial feature extraction step. Crucially, all three necessary read-out blocks are extracted simultaneously during a single forward pass. 

For $N = 500$, the complete evaluation pipeline executed in approximately $7.87$\,s on a single NVIDIA A6000 GPU. In stark contrast, calculating traditional manifold precision and recall at the exact same sample size required $80.92$\,s. Memory overhead is also exceptionally minimal. Storing the single-precision embeddings of both datasets across all three blocks requires only about $4.6$\,MB of space.

Beyond the pre-specified extraction depths, the framework relies on three fixed hyperparameters. We use $k = 5$ nearest neighbors for the coverage axis, five bandwidth scales for the kernel (Equation~\ref{eq:bandwidths}), and $B = 500$ random permutations for statistical inference. We recommend a sample size of $N \ge 100$ for rapid monitoring during model training. For formal benchmarking and reporting, researchers should use $N = 500$. Finally, this framework easily generalizes to other radiology modalities without requiring retraining or modality-specific configurations. The only strict requirement is that the real reference images must be sampled in a subject-diverse manner, exactly as specified in Section~\ref{sec:method_reference}.

\section{Experiments and Results}
\label{sec:results}

\subsection{Experimental Setup}
\label{sec:setup}

\paragraph{Datasets}
We utilized the BraTS 2021 dataset \citep{menze2014multimodal, bakas2018identifying, baid2021rsna} as our primary data source. Specifically, we extracted axial slices 55 through 85 from the FLAIR volume of each of the 1,251 available subjects. This extraction yielded 31 slices per subject, culminating in a total of 38,781 individual slices, each with a spatial resolution of $240 \times 240$ pixels. We applied a quality control filter requiring a minimum of 5\% non-zero pixels per slice; however, this constraint ultimately resulted in no slice removals. For preprocessing, we clipped pixel intensities at the 1st and 99th percentiles, computed exclusively over the non-zero voxels. Subsequently, we normalized these intensities to the $[0, 1]$ range.

For the lesion-specificity experiments, we employed the lower-grade glioma (LGG) dataset introduced by \cite{buda2019association}. This dataset contains preoperative MRI of 110 patients from The Cancer Genome Atlas lower-grade glioma collection, with 20 to 88 slices per patient stored as standard 8-bit images, together with manual segmentations of FLAIR abnormalities verified by a board-eligible radiologist. We used a subset of 600 image-mask pairs from this dataset.
These expert masks provided the precise ground truth necessary for our localized evaluations.

\paragraph{Generative Models}
We generated the in-domain dataset using an unconditional Denoising Diffusion Probabilistic Model (DDPM) \cite{ho2020denoising}. We initialized this model using a DDPM pretrained on the CelebA-HQ dataset at $256 \times 256$ pixel resolution \cite{ho2020denoising, karras2018progressive}. We then adapted it for single-channel input and fine-tuned it for 300 epochs across all 38,781 BraTS slices from the 1,251 available subjects. For training, we employed the $\epsilon$-prediction mean squared error objective alongside a linear noise schedule, with $\beta$ increasing from $10^{-4}$ to $0.02$ over $T = 1000$ timesteps. We utilized the AdamW optimizer \cite{loshchilov2019adamw} with a constant learning rate of $10^{-5}$, a batch size of 16, and mixed-precision arithmetic, notably excluding an exponential moving average of the weights. We generated $256 \times 256$ pixel samples using the complete 1000-step reverse process and retained all outputs without post-hoc selection. Because we trained the DDPM on all available subjects, the reference and held-out real sets inherently represent subsets of its training data.

We produced the remaining comparison sets using publicly available pretrained models. Specifically, we generated the weaker same-modality set and the cross-modality set using 3D wavelet diffusion models (WDM-3D) \cite{friedrich2024wdm}. These models were independently trained on the BraTS 2023 and LIDC-IDRI \cite{armato2011lidc} datasets at a resolution of $128^3$ voxels for 1.2 million iterations using 1000 linear diffusion steps. The BraTS model was trained on T1-weighted volumes, whereas our reference set consists of FLAIR slices; the weaker same-modality set therefore differs from the reference in MR sequence as well as in generator quality. For each WDM-3D model, we generated 20 full volumes and extracted 51 axial slices from the central 30 to 70\% of each volume. This process yielded 1,020 slices per model. We then bicubically upsampled these slices to $256 \times 256$ pixels and randomly selected 500 slices to form each evaluation set. 

To construct the far-domain set, we utilized a pretrained DDPM \citep{gs23retinal} trained on the moderate diabetic retinopathy subset of the Kaggle Retinal Fundus Images dataset. We generated 500 retinal fundus images using 1000 sampling steps at $128 \times 128$ pixel resolution and subsequently converted them to grayscale. Because both the cross-modality and far-domain sets are entirely synthetic, they inherently reflect shifts in modality or domain convoluted with the specific generative artifacts of their respective models.

\paragraph{Comparison Metrics}
We compared our $M^3$-Score against several established baselines. These included FID and KID computed on InceptionV3 features, as well as CMMD computed on CLIP ViT-L/14 features following the implementation by \cite{jayasumana2024cmmd}. Furthermore, we evaluated standard pixel-wise and perceptual metrics (SSIM, PSNR, MS-SSIM, LPIPS) alongside manifold-based $\alpha$-precision and $\beta$-recall using $k = 5$.

\paragraph{Protocol}
Unless otherwise stated, each evaluation dataset contained exactly $N = 500$ images. We sampled all real sets strictly across subjects as detailed in Section~\ref{sec:method_reference}. Our standard reference set comprised 500 BraTS slices drawn from 500 distinct patients. Prior to encoding, we scaled all image intensities to the $[0, 1]$ range and resized every image to $224 \times 224$ pixels. We rigorously maintained this exact configuration across all experiments to ensure fair comparisons.

\paragraph{Reported Quantities}
Throughout the results, $\rho$ denotes the Spearman rank correlation between a given perturbation level and the resulting metric value. We use $\tau$ to represent the CKA redundancy threshold, which is further detailed in Section~\ref{sec:results_depth}. The coefficient of variation (CV) indicates the standard deviation across repeated draws divided by the mean, reported as a percentage. The variable $Z$ refers to the null-standardized separation defined in Section~\ref{sec:method_mmd}. For the manifold metrics, $\alpha$-precision represents the fraction of generated samples falling within the $k$-NN manifold of the real set, while $\beta$-recall measures the fraction of real samples covered by the generated $k$-NN manifold \citep{kynkaanniemi2019pr, alaa2022faithful}. Finally, $\Delta$ signifies the difference between two comparative rows in a given table.

\subsection{Three-Axis Evaluation Across Comparison Sets}
\label{sec:results_canonical}

\begin{figure}[pos=htbp]
\centering
\includegraphics[width=\textwidth]{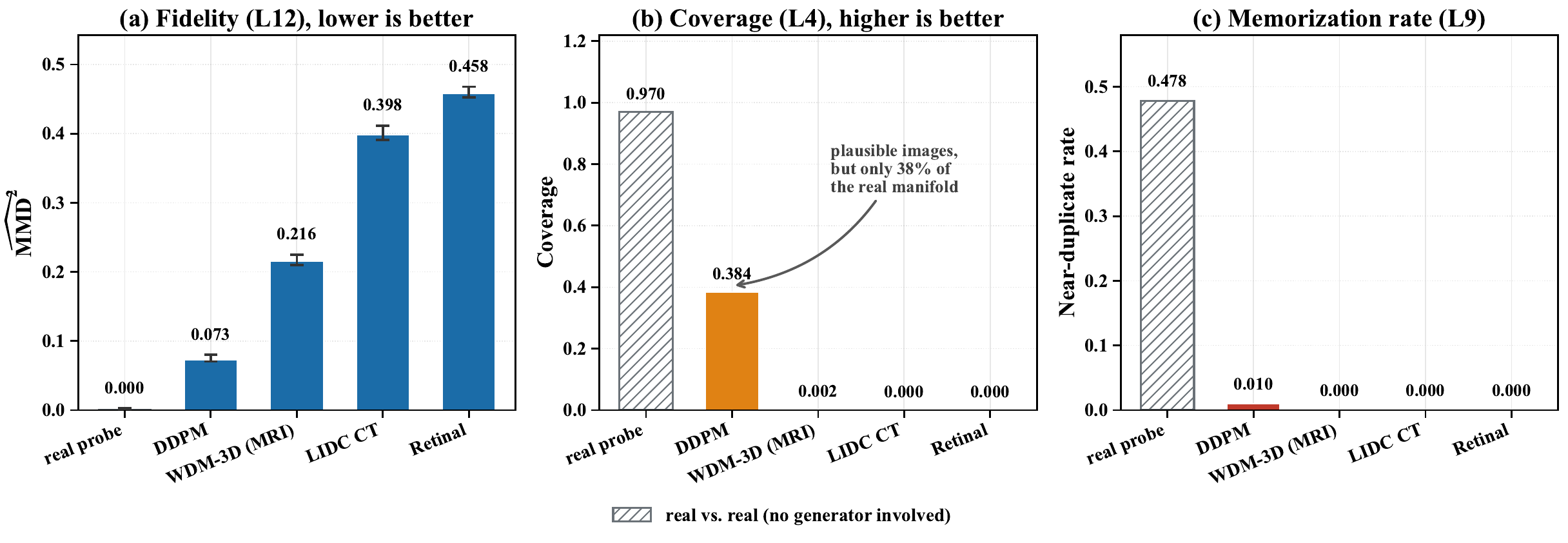}
\caption{%
Three-axis evaluation of five comparison sets of increasing expected severity against a reference of $N=500$ BraTS slices from $500$ subjects. The leftmost (hatched) bar denotes the subject-disjoint held-out real set. \textbf{(a)} Fidelity with bootstrap $95\%$ intervals; the interval of the held-out real set is not interpreted (see Table~\ref{tab:canonical}). \textbf{(b)} Coverage: the DDPM has the lowest fidelity distance among the generated sets but covers approximately $38\%$ of the real distribution. \textbf{(c)} Memorization: rates are close to zero for all generated sets and high for the held-out real set.}
\label{fig:three_axis}
\end{figure}

\begin{figure}[pos=htbp]
\centering
\includegraphics[width=\textwidth]{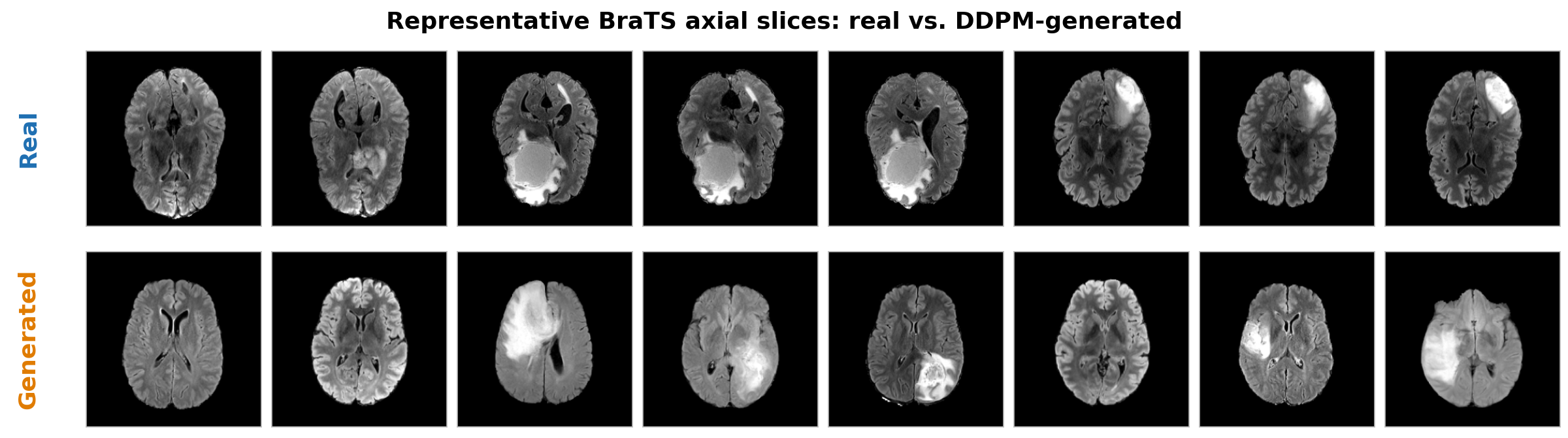}
\caption{Representative axial BraTS slices: real (top) and generated by the unconditional DDPM (bottom). Both sets exhibit plausible tissue boundaries, ventricles, and FLAIR-hyperintense lesions, and the two sets are difficult to distinguish by visual inspection. Images are shown at the $224\times224$ resolution used as encoder input.}
\label{fig:samples}
\end{figure}

Table~\ref{tab:canonical} and Fig.~\ref{fig:three_axis} report the three axes for five comparison sets ordered by expected severity: a subject-disjoint held-out real set, the DDPM, WDM-3D, synthetic chest CT, and synthetic retinal fundus images (Section~\ref{sec:setup}). All sets were evaluated against the standard reference. Representative real and DDPM-generated slices are shown in Fig.~\ref{fig:samples}.

\paragraph{Held-out real set}
For the held-out real set, the fidelity estimate was $0$ (permutation $p = 1$, $Z = -0.6$), indicating no detectable distributional shift between two independent groups of subjects. When the reference was instead drawn in filename order, the two real sets originated from different acquisition sites and a shift was detected, which illustrates the influence of reference construction. The coverage of the held-out real set was $0.970$. Its memorization rate was $0.478$, consistent with the value of approximately $0.5$ expected for two samples from the same distribution, since $\theta$ is the median real-to-real nearest-neighbor distance.

\paragraph{Fidelity and coverage}
Fidelity increased monotonically with expected severity ($0 < 0.073 < 0.216 < 0.398 < 0.459$; Spearman $\rho = 1.00$), and coverage did not increase ($0.970$, $0.384$, $0.002$, $0$, and $0$). The two axes nonetheless captured different properties. The DDPM obtained the lowest fidelity distance among the generated sets ($0.073$) but covered only $38.4\%$ of the real distribution. For WDM-3D, the fidelity distance was approximately three times that of the DDPM, whereas coverage was lower by a factor of almost $200$ ($0.002$).

\paragraph{Memorization}
The memorization rate at L9 was $0.010$ for the DDPM and $0$ for the remaining generated and out-of-domain sets, compared with $0.478$ for the held-out real set. Because the reference slices are part of the DDPM training data, these results indicate that almost none of the DDPM samples are near-duplicates of the reference slices.

\paragraph{Statistical inference}
All generated and out-of-domain sets were rejected at the minimum attainable $p$-value ($\hat{p} = 0.002$), whereas the held-out real set was not rejected. Because the null hypothesis specifies identical distributions, rejection indicates a detectable difference rather than a level of quality, and the $p$-values do not differentiate among the generated sets. The bootstrap interval of the held-out real set, $[0.0010, 0.0026]$, excludes the observed value of $0$. This reflects the inconsistency of the ordinary bootstrap for degenerate U-statistics under $H_0$~\citep{arcones1992bootstrap}, and this interval is therefore not interpreted. For the DDPM, the bootstrap interval $[0.0705, 0.0801]$ has a width of $0.0096$, and the observed statistic lies $342$ null standard deviations above the null mean.

\begin{table}[t]
\centering
\caption{
Three-axis evaluation of five comparison sets against a reference of $N=500$ BraTS slices from $500$ subjects. Rows are ordered by expected severity. Fidelity is reported with a permutation test ($B=500$) and a bootstrap $95\%$ CI ($500$ resamples). $^{\ddagger}$Not interpreted, because the ordinary bootstrap is inconsistent for a degenerate U-statistic under $H_0$.}
\label{tab:canonical}
\footnotesize
\setlength{\tabcolsep}{3pt}
\begin{tabular}{llcccccc}
\toprule
& & \multicolumn{4}{c}{Fidelity (L12)} & Memorization (L9) & Coverage (L4) \\
\cmidrule(lr){3-6}\cmidrule(lr){7-7}\cmidrule(lr){8-8}
Comparison set & Content & $\widehat{\mathrm{MMD}}^2$ & 95\% CI & $p$ & $Z$
 & Rate & Cov. \\
\midrule
Held-out real  & real vs.\ real & $0$ & $[0.0010, 0.0026]^{\ddagger}$ & $1$ & $-0.6$   & $0.478$ & $0.970$ \\
\midrule
DDPM           & in-domain diffusion            & $0.0732$ & $[0.0705, 0.0801]$ & $0.002$ & $342.4$ & $0.010$ & $0.384$ \\
WDM-3D (MRI)   & weaker same-modality generator & $0.2159$ & $[0.2102, 0.2250]$ & $0.002$ & $696.9$ & $0$ & $0.002$ \\
LIDC CT        & cross-modality shift           & $0.3984$ & $[0.3909, 0.4117]$ & $0.002$ & $718.4$ & $0$ & $0$ \\
Retinal        & far-domain shift               & $0.4585$ & $[0.4525, 0.4678]$ & $0.002$ & $805.3$ & $0$ & $0$ \\
\bottomrule
\end{tabular}
\end{table}

\subsection{Response to Noise and Blur}
\label{sec:results_noise}

An evaluation metric is expected to respond monotonically to increasing image degradation. Gaussian noise ($\sigma \in \{0, 0.01, 0.02, 0.05, 0.1, 0.2, 0.5\}$) and Gaussian blur (radius $r \in \{0, 1, 2, 3, 4\}$ pixels) were applied to $N = 500$ real BraTS images, and ten metrics were computed between the clean and degraded sets. Table~\ref{tab:noise_spearman} reports the Spearman correlation between degradation level and metric value.

\begin{table}[t]
\centering
\caption{%
  Spearman rank correlation $\rho$ between degradation level and
  metric value (seven noise levels; five blur radii).
  $\uparrow$: the metric is expected to increase with degradation;
  $\downarrow$: the metric is expected to decrease.
}
\label{tab:noise_spearman}
\setlength{\tabcolsep}{6pt}
\begin{tabular}{lcccc}
\toprule
\multirow{2}{*}{Metric} &
\multirow{2}{*}{Expected} &
\multicolumn{2}{c}{Spearman $\rho$} \\
\cmidrule(lr){3-4}
& & Noise & Blur \\
\midrule
M3-Score              & $\uparrow$   & $1.000$          & $1.000$ \\
FID (Inception)       & $\uparrow$   & $1.000$          & $1.000$ \\
KID (Inception)       & $\uparrow$   & $1.000$          & $1.000$ \\
CMMD (CLIP)           & $\uparrow$   & $1.000$          & $1.000$ \\
SSIM                  & $\downarrow$ & $-1.000$         & $-1.000$ \\
PSNR                  & $\downarrow$ & $-1.000$         & $-1.000$ \\
MS-SSIM               & $\downarrow$ & $-1.000$         & $-1.000$ \\
LPIPS                 & $\uparrow$   & $1.000$          & $1.000$ \\
$\alpha$-Precision    & $\downarrow$ & $-0.982$         & $-0.975$ \\
$\beta$-Recall        & $\downarrow$ & $-0.982$         & $-0.975$ \\
\bottomrule
\end{tabular}
\end{table}

\begin{figure}[pos=htbp]
\centering
\includegraphics[width=0.85\linewidth]{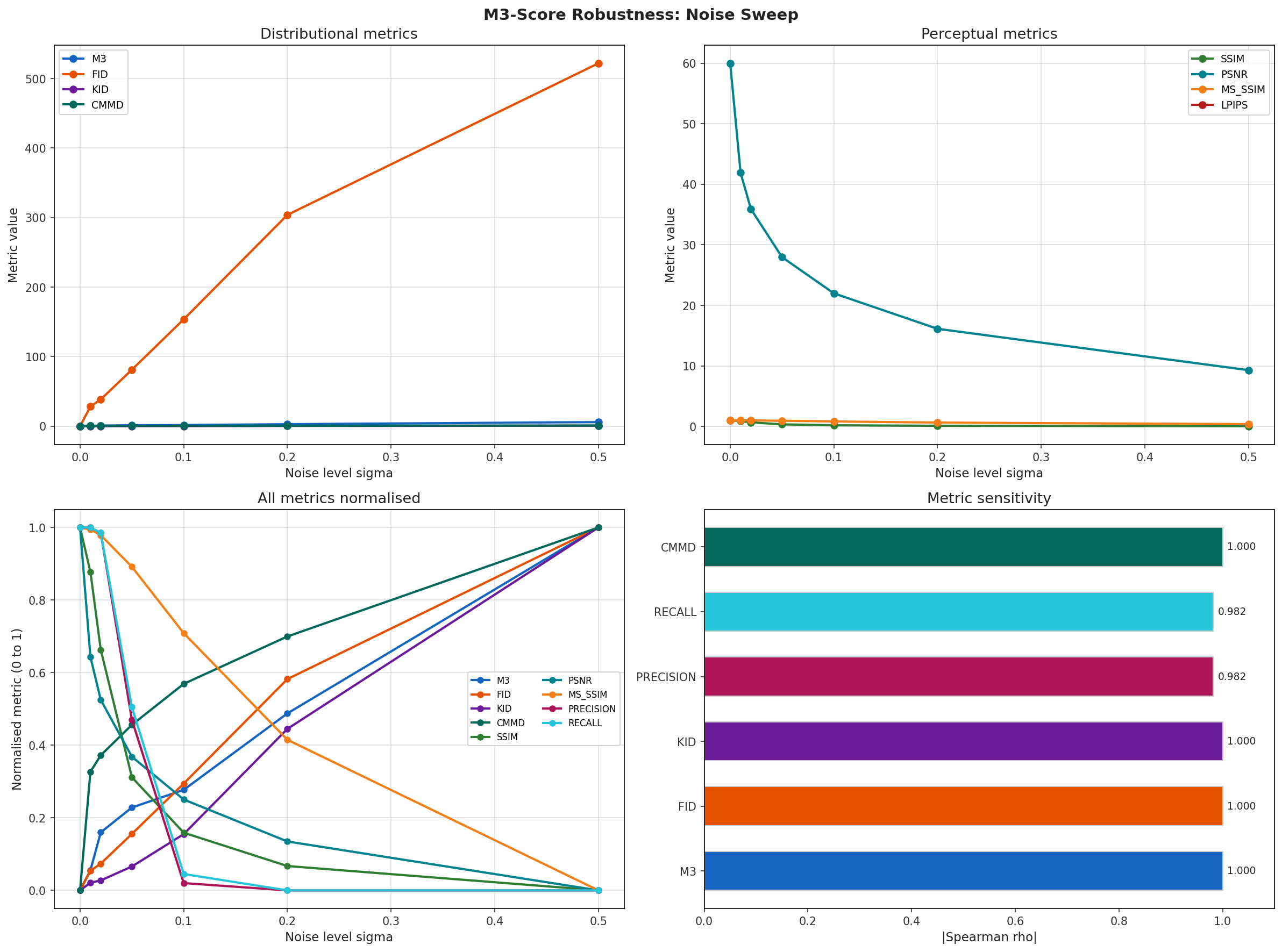}
\caption{%
\textbf{Gaussian noise sweep} ($\sigma\in[0,0.5]$; $N=500$ BraTS images). Top left: distributional metrics; top right: perceptual metrics; bottom left: all metrics rescaled to $[0,1]$; bottom right: $|\mathrm{Spearman}\ \rho|$ for M3, FID, KID, CMMD, $\alpha$-Precision, and $\beta$-Recall (labeled PRECISION and RECALL).
}

\label{fig:noise_robustness}
\end{figure}

\begin{figure}[pos=htbp]
\centering
\includegraphics[width=0.85\linewidth]{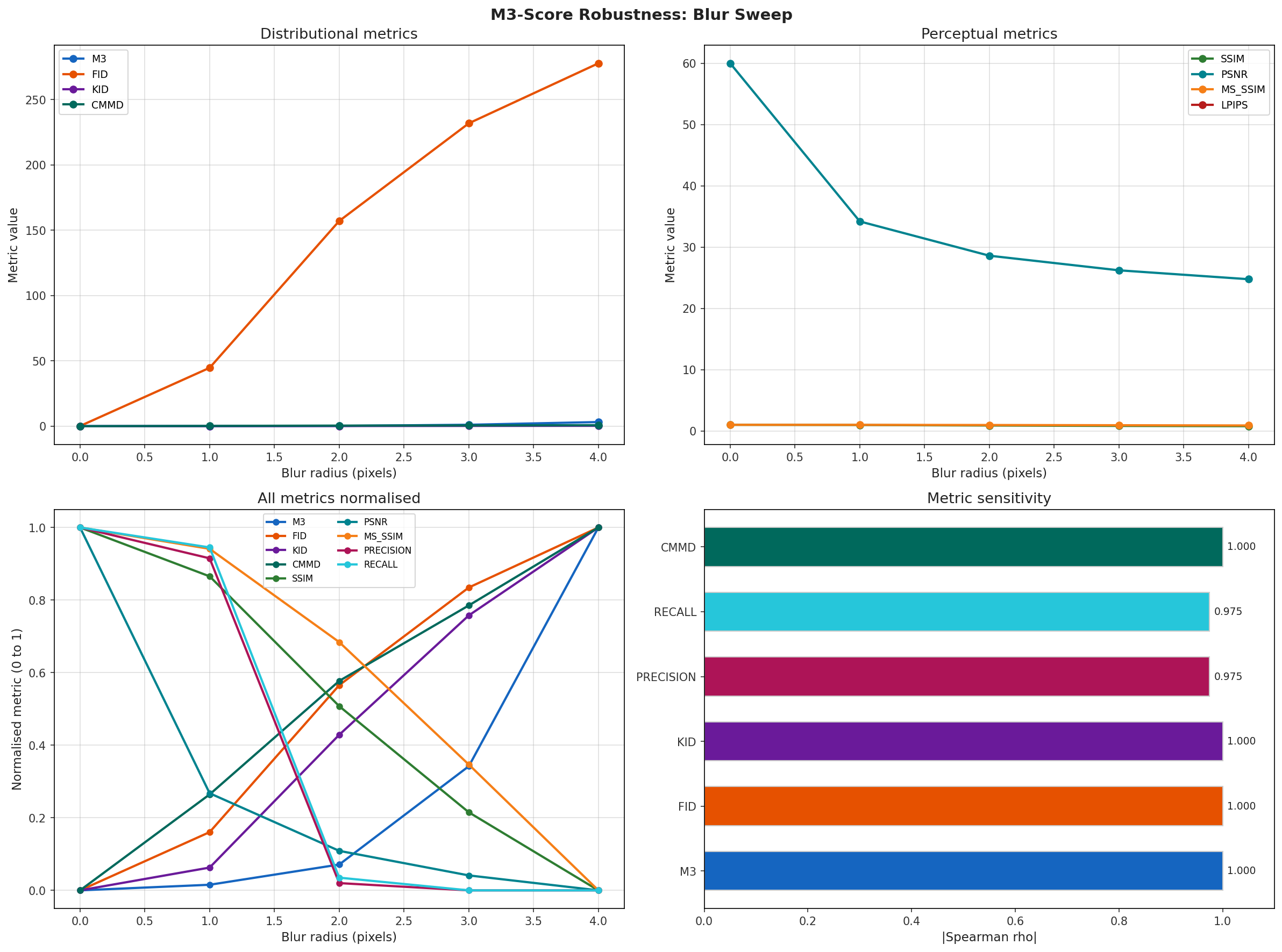}
\caption{%
\textbf{Gaussian blur sweep} ($r\in\{0,1,2,3,4\}$ pixels; $N=500$ BraTS images). Top left: distributional metrics; top right: perceptual metrics; bottom left: all metrics rescaled to $[0,1]$; bottom right: $|\mathrm{Spearman}\ \rho|$ for M3, FID, KID, CMMD, $\alpha$-Precision, and $\beta$-Recall (labeled PRECISION and RECALL). $\alpha$-Precision decreases to $0.006$ at $r=2$, indicating that the BraTS manifold is strongly dependent on fine texture.
}

\label{fig:blur_robustness}
\end{figure}

\paragraph{Noise}
For $\sigma = 0$, the clean set is compared with itself, and the M3 fidelity estimate was $0$. All metrics except $\alpha$-Precision and $\beta$-Recall ($|\rho| = 0.982$) were perfectly rank-correlated with the noise level ($|\rho| = 1$; Fig.~\ref{fig:noise_robustness}). At $\sigma = 0.5$, the fidelity MMD$^2$ reached $0.607$, FID $521.96$, and KID $0.868$, whereas SSIM decreased to $0.031$ and PSNR to $9.29$\,dB. These values are unnormalized; the bottom-left panels of Figs.~\ref{fig:noise_robustness} and~\ref{fig:blur_robustness} present all metrics rescaled to $[0,1]$.

\paragraph{Blur}
Under Gaussian blur, M3 also attained $\rho = 1$, as did FID, KID, and CMMD, and its value increased strictly across all five radii (Fig.~\ref{fig:blur_robustness}). $\alpha$-Precision indicated that blur displaces images from the real manifold to a greater extent than noise at comparable image quality. At blur radius $r = 2$ and at noise level $\sigma = 0.05$, the PSNR was approximately $28$\,dB; nevertheless, $\alpha$-Precision decreased to $0.006$ under blur but remained $0.384$ under noise. This result suggests that the BraTS image manifold is characterized largely by fine texture, which blur removes and additive noise partially preserves.

\subsection{Feature-Space Comparison for Out-of-Distribution Discrimination}
\label{sec:results_ood}

This section compares the RadioDINO-s16, InceptionV3, and CLIP feature spaces in their ability to separate real brain MRI from generated and OOD images.

\paragraph{Agreement with an anomaly detector}
Agreement between the per-image score of Eq.~\eqref{eq:perim}, computed in each feature space, and an Isolation Forest anomaly detector~\citep{liu2008isolation} fitted on the real reference was first examined. Because the detector must be fitted in a particular feature space, detectors were fitted in both the RadioDINO-s16 and the InceptionV3 spaces (Table~\ref{tab:ood_spearman}). Each distance correlated strongly with the detector fitted in the same space ($\rho = +0.736$ for the RadioDINO-s16 score with the RadioDINO-s16 detector and $+0.649$ for the InceptionV3 score with the InceptionV3 detector) and weakly with the detector fitted in the other space ($+0.144$ and $+0.155$, respectively). The same-space correlations exceeded the cross-space correlations by a factor of four to five, with non-overlapping confidence intervals, whereas pixel MSE correlated similarly with both detectors ($+0.288$ and $+0.364$). Detector-based agreement therefore reflects the shared representation rather than anomaly-detection ability and is not suitable for comparing feature spaces. Per-image discrimination was consequently assessed against image provenance.

\begin{table}[t]
\centering
\caption{%
  Spearman correlation between per-image distances and Isolation Forest anomaly scores, with the detector fitted in each feature space ($N = 500$ generated images; $95\%$ CI by Fisher $z$-transformation). Bold entries denote cases in which the distance and the detector share a feature space.
}
\label{tab:ood_spearman}
\setlength{\tabcolsep}{4pt}
\small
\begin{tabular}{lcc}
\toprule
& \multicolumn{2}{c}{Isolation Forest fitted in} \\
\cmidrule(lr){2-3}
Per-image distance & RadioDINO-s16 & InceptionV3 \\
\midrule
RadioDINO-s16 score (M3)      & $\mathbf{+0.736}\,[+0.69,+0.77]$ & $+0.144\,[+0.06,+0.23]$ \\
InceptionV3 score             & $+0.155\,[+0.07,+0.24]$ & $\mathbf{+0.649}\,[+0.60,+0.70]$ \\
Pixel MSE                     & $+0.288\,[+0.21,+0.37]$ & $+0.364\,[+0.29,+0.44]$ \\
\bottomrule
\end{tabular}
\end{table}

\begin{figure}[pos=htbp]
\centering
\includegraphics[width=\textwidth]{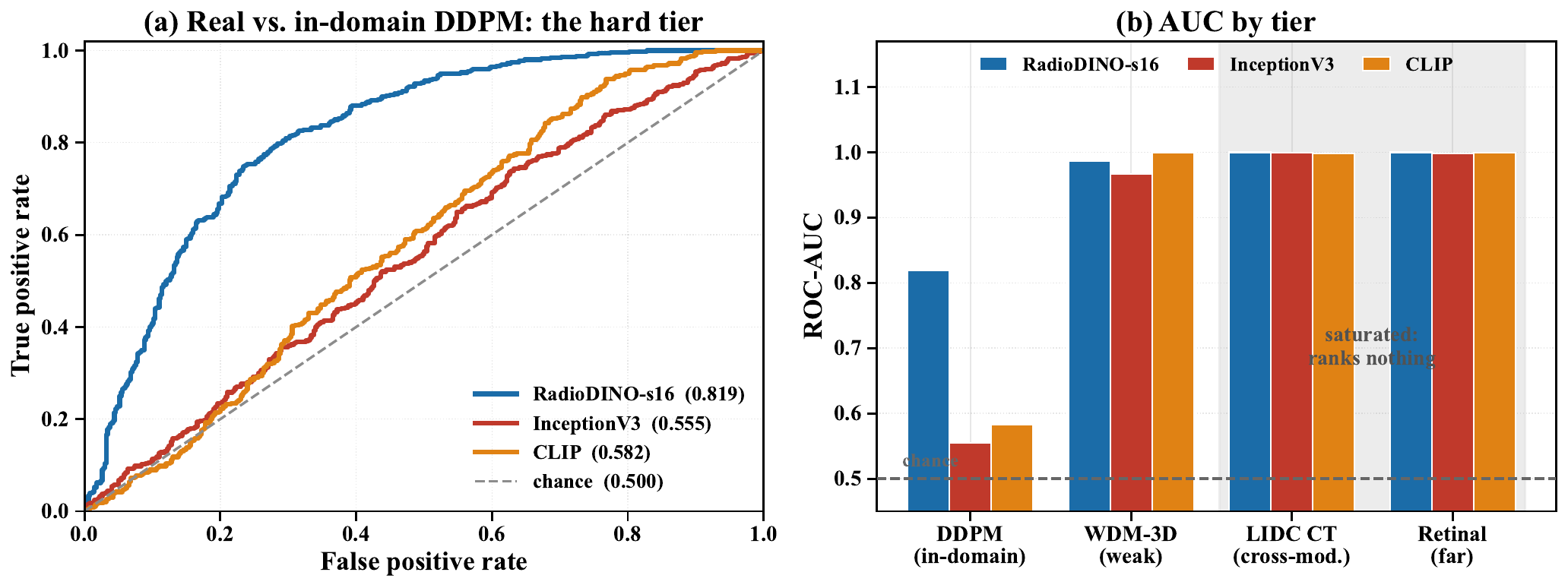}
\caption{
Per-image OOD discrimination using a centroid-distance score in three feature spaces. The positive class is a subject-disjoint held-out real set. \textbf{(a)} ROC curves for real images versus DDPM samples: RadioDINO-s16 attains AUC $=0.819$, compared with $0.555$ for InceptionV3 (FID) and $0.582$ for CLIP (CMMD). \textbf{(b)} AUC for each comparison set. The cross-modality and far-domain sets (shaded) saturate at $\geq0.998$ in all feature spaces.
}

\label{fig:ood}
\end{figure}

\paragraph{Per-image discrimination}
Using the per-image score of Eq.~\eqref{eq:perim} in each of the three feature spaces, the ROC-AUC was computed between a subject-disjoint held-out real set and each comparison set (Table~\ref{tab:ood_metric}, Fig.~\ref{fig:ood}). For the cross-modality CT and retinal sets, all feature spaces attained an AUC of at least $0.998$; these sets therefore do not discriminate between feature spaces, and results are reported separately for each set rather than pooled.

The feature spaces differed substantially on the in-domain task of separating real brain MRI from DDPM samples. RadioDINO-s16 attained a ROC-AUC of $0.819$, whereas InceptionV3 ($0.555$) and CLIP ($0.582$) performed close to chance. Expressed as the margin above chance ($\mathrm{AUC} - 0.5$), the margin of RadioDINO-s16 was approximately $5.8$ times that of InceptionV3 and $3.9$ times that of CLIP. For the WDM-3D set, CLIP ($0.999$) attained a slightly higher AUC than RadioDINO-s16 ($0.986$). The advantage of the radiology-pretrained representation was therefore concentrated on the in-domain generator, for which discrimination is most difficult.

\begin{table}[t]
\centering
\caption{%
  Per-image OOD discrimination by feature space. ROC-AUC for separating a subject-disjoint held-out real set from each comparison set, using the distance to the real-reference centroid on normalized features. $N = 500$ per set; the reference contains $500$ slices from $500$ subjects.}
\label{tab:ood_metric}
\setlength{\tabcolsep}{4pt}
\small
\begin{tabular}{lcccc}
\toprule
& \multicolumn{2}{c}{Discriminative} & \multicolumn{2}{c}{Saturated} \\
\cmidrule(lr){2-3}\cmidrule(lr){4-5}
Feature space & DDPM & WDM-3D & LIDC CT & Retinal \\
\midrule
\textbf{RadioDINO-s16 L12} & $\mathbf{0.819}$ & $0.986$ & $1.000$ & $1.000$ \\
InceptionV3 2048-d (FID)   & $0.555$ & $0.967$ & $1.000$ & $0.999$ \\
CLIP ViT-L/14 (CMMD)       & $0.582$ & $\mathbf{0.999}$ & $0.998$ & $0.999$ \\
\bottomrule
\end{tabular}
\end{table}

\begin{figure}[pos=htbp]
\centering
\includegraphics[width=\textwidth]{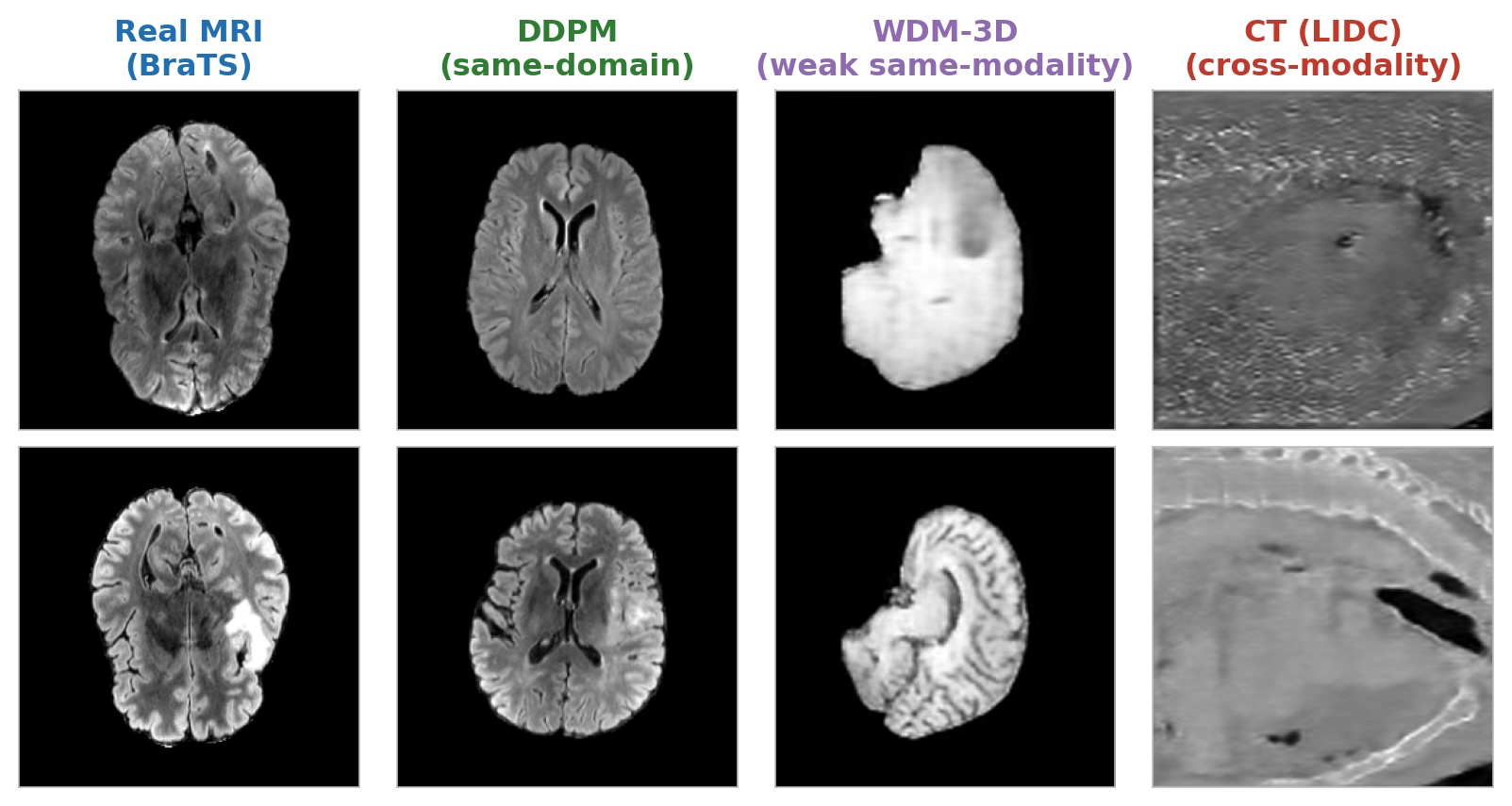}
\caption{Examples of the comparison sets evaluated against real BraTS MRI, with two examples in each column. From left: real MRI; DDPM samples; WDM-3D samples, which are of visibly lower quality but remain brain MRI; and chest CT slices generated by the WDM-3D model trained on LIDC-IDRI, which differ in both modality and anatomy. M3 and FID place the CT set furthest from real brain MRI, whereas CMMD places it closer than the WDM-3D set (Table~\ref{tab:cmmd_fail}).}
\label{fig:tiers}
\end{figure}

\paragraph{Ordering of comparison sets}
The ability of each metric to order comparison sets was evaluated for three sets: the DDPM, WDM-3D, and the synthetic chest CT set (Fig.~\ref{fig:tiers}). The expected order is DDPM $<$ WDM-3D $<$ CT, because a different modality and anatomy constitutes the largest shift. The WDM-3D set is T1-weighted, whereas the reference is FLAIR (Section~\ref{sec:setup}), so its distance reflects an MR sequence shift in addition to lower generator quality. CMMD was computed with CLIP ViT-L/14 image features and an unbiased Gaussian RBF MMD$^2$ with the median-heuristic bandwidth computed on the pooled sample, following \citet{jayasumana2024cmmd}.

The M3 fidelity axis reproduced the expected order with non-overlapping intervals ($0.073 < 0.216 < 0.398$; Table~\ref{tab:cmmd_fail}), and FID also reproduced it ($68.4 < 173.4 < 319.9$). CMMD reversed the last two sets, assigning the cross-modality CT set ($0.664\,[0.652, 0.682]$) a smaller distance than the WDM-3D set ($0.737\,[0.719, 0.754]$). The two intervals did not overlap. The difference between the CT and WDM-3D sets was $-0.073$ for CMMD and $+0.182$ for M3. Because FID preserved the expected order, the reversal is attributable to the CLIP representation rather than to natural-image backbones in general.

\begin{table}[t]
\centering
\caption{%
Distances from real BraTS images ($N=500$) for three comparison sets, with bootstrap $95\%$ CIs for M3 and CMMD ($500$ resamples). Higher values indicate a larger distance, and the expected order is DDPM $<$ WDM-3D $<$ CT. M3 and FID follow this order, whereas CMMD reverses the last two sets. FID is reported as a point estimate.
}

\label{tab:cmmd_fail}

\setlength{\tabcolsep}{3pt}
\small
{%
\begin{tabular}{lccc}
\toprule
Comparison vs.\ BraTS real & M3 (fidelity) [95\% CI] & FID & CMMD [95\% CI] \\
\midrule
DDPM (in-domain)                 & $0.073\,[0.071,0.080]$ & $68.4$  & $0.104\,[0.102,0.111]$ \\
WDM-3D (weaker same-modality)    & $0.216\,[0.210,0.225]$ & $173.4$ & $0.737\,[0.719,0.754]$ \\
LIDC CT (cross-modality)         & $\mathbf{0.398}\,[0.391,0.412]$ & $319.9$ & $0.664\,[0.652,0.682]$ \\
\midrule
$\Delta$ (CT $-$ WDM-3D)         & $\mathbf{+0.182}$ & $+146.5$ & $-0.073$ \\
\bottomrule
\end{tabular}}
\end{table}

\paragraph{Cross-modality and far-domain sets}
The M3 fidelity distance to brain MRI was larger for the synthetic retinal fundus images ($0.459$) than for the synthetic chest CT images ($0.398$; Table~\ref{tab:canonical}). This ordering is consistent with the pretraining domain of RadioDINO, which includes CT and MRI but not fundus photography. For images outside the radiology domain, M3 values reflect distances in a radiology-specific representation and should not be interpreted as general measures of visual distance.

\subsection{Validation of the Memorization Axis}
\label{sec:results_axes}

The memorization axis was validated by replacing a known fraction of the generated set with near-duplicates of real images, obtained by adding Gaussian jitter ($\sigma = 5\times10^{-4}$) to real images, at injection rates of $\{0, 0.1, 0.2, 0.3, 0.5\}$. The standard reference and the DDPM set were used.

The L9 memorization rate followed the injected fraction ($0.010$, $0.108$, $0.210$, $0.308$, and $0.506$) with a mean absolute error of $0.008$ (Table~\ref{tab:axes}). The rate of $0.010$ at zero injection corresponds to the near-duplicate rate of the DDPM itself (Table~\ref{tab:canonical}). Over the same range, the fidelity MMD$^2$ decreased from $0.073$ to $0.017$ and coverage increased from $0.384$ to $0.982$. Because the injected images are derived from real data, this perturbation moves the generated set toward the real distribution on all three axes simultaneously, and the axes are correlated under this perturbation.

\begin{table}[t]
\centering
\caption{%
  Near-duplicate injection on the standard reference ($N = 500$ from $500$ subjects). A known fraction of the DDPM set is replaced by jittered copies of real images. The memorization rate follows the injected fraction (mean absolute error $0.008$), while fidelity decreases and coverage increases.
}
\label{tab:axes}
\begin{tabular}{lccc}
\toprule
Injection rate & Mem.\ rate (L9) & Fidelity MMD$^2$ (L12) & Coverage (L4) \\
\midrule
$0$   & $0.010$ & $0.0732$ & $0.384$ \\
$0.1$ & $0.108$ & $0.0597$ & $0.674$ \\
$0.2$ & $0.210$ & $0.0470$ & $0.810$ \\
$0.3$ & $0.308$ & $0.0346$ & $0.906$ \\
$0.5$ & $0.506$ & $0.0173$ & $0.982$ \\
\bottomrule
\end{tabular}
\end{table}

\subsection{Validation of the Coverage Axis}
\label{sec:results_coverage}

\begin{figure}[pos=htbp]
\centering
\includegraphics[width=\linewidth]{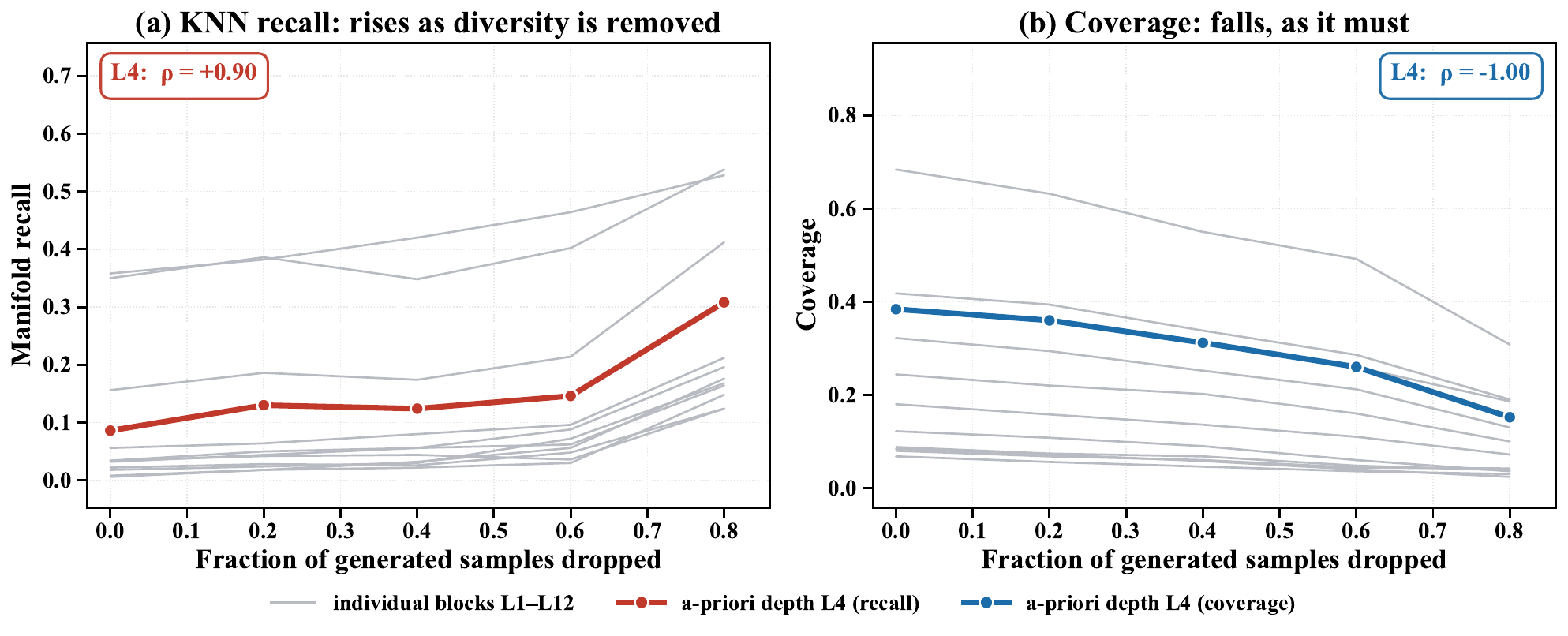}
\caption{%
Mode-drop response at all twelve encoder blocks (gray lines), with the pre-specified depth L4 highlighted. \textbf{(a)} $k$-NN manifold recall increases as diversity is removed, because its hyperspheres are fitted to the subsampled generated set. \textbf{(b)} Real-radius coverage decreases monotonically at every depth.}

\label{fig:coverage}
\end{figure}

The coverage axis was validated with a nested mode-drop experiment. A single random ordering of the generated set was drawn, and each level retained a prefix of this ordering, so that samples removed at one level remained removed at all higher levels. A nested design was used because a generator that loses modes does not resample its outputs; independent sampling at each level would confound sample removal with resampling variation. The drop fractions were $\{0, 0.2, 0.4, 0.6, 0.8\}$, with $N = 500$ real slices from $500$ subjects and $k = 5$. The acceptance criterion, specified before the experiment was conducted, required at L4 a Spearman correlation of at most $-0.9$ between drop fraction and the statistic and a range of at least $0.05$ across the levels. The range condition excludes statistics that are monotone but remain close to $0$ or $1$. Both estimators described in Section~\ref{sec:method_coverage} were evaluated at all twelve blocks (Table~\ref{tab:coverage}, Fig.~\ref{fig:coverage}).

\paragraph{$k$-NN manifold recall}
Recall increased with the drop fraction at all twelve blocks, with $\rho$ between $+0.70$ and $+1.00$. At L4, recall increased from $0.086$ to $0.308$ as $80\%$ of the generated set was removed; at L9, it increased from $0.008$ to $0.148$. The consistency of this behavior across blocks with very different representations indicates that it arises from the estimator rather than from the features, specifically from the enlargement of neighborhood radii described in Section~\ref{sec:method_coverage}.

\paragraph{Real-radius coverage}
Coverage as defined in Eq.~\eqref{eq:coverage} decreased monotonically at all twelve blocks ($\rho = -1.00$). At L4, it decreased from $0.384$ to $0.360$, $0.312$, $0.260$, and $0.152$ across the levels, a range of $0.232$, which satisfies the acceptance criterion. The range was largest at L1 ($0.376$), remained above $0.19$ from L2 to L5, and reached its minimum of $0.038$ at L10 and L11, where the range condition was not satisfied. This pattern is consistent with the encoding of spatial and structural variation in early blocks.

\paragraph{Implications for $k$-NN-based metrics}
$k$-NN precision and recall are widely used to diagnose generative models, and their neighborhood radii are estimated on the sets being compared. Protocols that subsample one of these sets, such as mode-drop analyses and sample-size studies, can therefore reverse the direction of the statistic. Estimating the radii on a fixed real reference avoids this effect.

\begin{table}[t]
\centering
\caption{Comparison of mode dropping responses between two coverage estimators across all twelve encoder blocks. We report the Spearman rank correlation ($\rho$) between the mode drop fraction ($\downarrow$) and the respective metric value. A theoretically valid coverage estimator must exhibit a strongly negative $\rho$. Range denotes the absolute difference between the maximum and minimum values across all perturbation levels. Our pre-specified extraction depth (L4) is highlighted in bold. The evaluation utilized $N = 500$ real slices from 500 distinct subjects in a nested design using $k = 5$ nearest neighbors.}
\label{tab:coverage}
\small
\begin{tabular}{ccccc}
\toprule
& \multicolumn{1}{c}{$k$-NN recall \citep{kynkaanniemi2019pr}}
& \multicolumn{2}{c}{Coverage \citep{naeem2020reliable}, Eq.~\eqref{eq:coverage}} & \\
\cmidrule(lr){2-2}\cmidrule(lr){3-4}
Block & $\rho(\downarrow, \text{recall})$ & $\rho(\downarrow, \text{cov.})$ & Range & Criterion \\
\midrule
L1  & $+1.00$ & $-1.00$ & $0.376$ & \checkmark \\
L2  & $+0.70$ & $-1.00$ & $0.196$ & \checkmark \\
L3  & $+0.90$ & $-1.00$ & $0.228$ & \checkmark \\
\textbf{L4}  & $\mathbf{+0.90}$ & $\mathbf{-1.00}$ & $\mathbf{0.232}$ & \textbf{\checkmark} \\
L5  & $+1.00$ & $-1.00$ & $0.192$ & \checkmark \\
L6  & $+1.00$ & $-1.00$ & $0.144$ & \checkmark \\
L7  & $+0.70$ & $-1.00$ & $0.108$ & \checkmark \\
L8  & $+0.90$ & $-1.00$ & $0.086$ & \checkmark \\
L9  & $+1.00$ & $-1.00$ & $0.060$ & \checkmark \\
L10 & $+1.00$ & $-1.00$ & $0.038$ & $\times$ \\
L11 & $+1.00$ & $-1.00$ & $0.038$ & $\times$ \\
L12 & $+1.00$ & $-1.00$ & $0.050$ & \checkmark \\
\midrule
\multicolumn{5}{l}{\footnotesize Recall: expected direction at 0/12 blocks.
  Coverage: expected direction at 12/12 blocks; both conditions met at 10/12 blocks.}\\
\bottomrule
\end{tabular}
\end{table}

\subsection{Sensitivity to Encoder Depth}
\label{sec:results_depth}

\paragraph{Memorization}
We repeated the near-duplicate injection experiment from Section~\ref{sec:results_axes} across all twelve encoder blocks. The metric successfully recovered the injected fraction at every depth. Consequently, this specific task does not strongly discriminate between layers, confirming that our reported memorization results remain robust and do not strictly depend on our pre-specified choice of L9.

\paragraph{Coverage}
As demonstrated in Table~\ref{tab:coverage}, the direction of the mode dropping response remained completely independent of depth for both evaluated estimators. The choice of depth exclusively affected the dynamic range of the real-radius coverage. This range naturally peaked at the shallower blocks of the network, which directly supports our selection of L4.

\paragraph{Fidelity and Layer Redundancy.}
We analyzed the redundancy of the encoder layers using greedy backward Centered Kernel Alignment (CKA) pruning, detailed in Appendix~\ref{app:cka}. We applied this technique to the real reference images using thresholds ($\tau$) ranging from 0.70 to 0.99, with results summarized in Table~\ref{tab:cka_ablation}. Notably, L12 was the only block retained at every single threshold, persisting from the most restrictive setting ($\tau = 0.70$, yielding only three retained layers) to the least restrictive setting ($\tau = 0.99$, retaining all layers). No other block exhibited this stability; for instance, L1 was pruned at $\tau = 0.70$, and the retention of mid-depth blocks fluctuated heavily based on the threshold.

Furthermore, the mean real-generated discrepancy over the retained layers, denoted as $S_{\text{RG}}$, varied minimally across thresholds (ranging from 0.087 to 0.092). This stability indicates that the extra layers retained at higher thresholds contribute very little additional discriminatory separation. The corresponding real-real discrepancy, $S_{\text{RR}}$, computed between two subject-disjoint real datasets, remained effectively zero at all thresholds (ranging from $2 \times 10^{-5}$ to $12 \times 10^{-5}$). Because $S_{\text{RR}}$ is so close to zero, ratio-based summaries such as $(S_{\text{RG}} - S_{\text{RR}})/S_{\text{RR}}$ become inherently numerically unstable under our rigorous subject-level sampling protocol. We therefore strictly avoid using them.

\begin{table}[t]
\centering
\caption{%
  CKA-based layer-redundancy analysis on the standard reference ($N = 500$ from $500$ subjects). $M$ is the number of retained layers. $S_\text{RG}$ is the mean MMD$^2$ between the real and generated sets over the retained layers, and $S_\text{RR}$ is the corresponding value between two subject-disjoint real sets. L12 (bold) is retained at every threshold.
}
\label{tab:cka_ablation}
\setlength{\tabcolsep}{4pt}
\small
\begin{tabular}{ccccc}
\toprule
$\tau$ & $M$ & Retained layers & $S_\text{RG}$ & $S_\text{RR}$ \\
\midrule
$0.70$ & $3$  & $\{2, 5, \mathbf{12}\}$                  & $0.0915$ & $1.2{\times}10^{-4}$ \\
$0.80$ & $4$  & $\{1, 2, 6, \mathbf{12}\}$               & $0.0866$ & $0.6{\times}10^{-4}$ \\
$0.85$ & $5$  & $\{1, 2, 4, 7, \mathbf{12}\}$            & $0.0907$ & $0.2{\times}10^{-4}$ \\
$0.90$ & $6$  & $\{1, 2, 4, 6, 8, \mathbf{12}\}$         & $0.0879$ & $0.4{\times}10^{-4}$ \\
$0.92$ & $6$  & $\{1, 2, 4, 6, 8, \mathbf{12}\}$         & $0.0879$ & $0.4{\times}10^{-4}$ \\
$0.95$ & $7$  & $\{1, 2, 4, 5, 7, 9, \mathbf{12}\}$      & $0.0900$ & $0.7{\times}10^{-4}$ \\
$0.97$ & $8$  & $\{1, 2, 4$--$7, 9, \mathbf{12}\}$       & $0.0890$ & $0.9{\times}10^{-4}$ \\
$0.99$ & $12$ & $\{1$--$\mathbf{12}\}$                   & $0.0874$ & $0.6{\times}10^{-4}$ \\
\bottomrule
\end{tabular}
\end{table}

\subsection{Lesion Specificity}
\label{sec:results_masking}

A radiology-pretrained representation should naturally respond more strongly to the degradation of diagnostically relevant tissue than to an equivalent degradation of healthy tissue. We explicitly examined this expected property using the expert FLAIR abnormality masks provided by the LGG dataset \citep{buda2019association}.

\paragraph{Design.}
For each image, we perturbed either the annotated tumor region (\textsc{lesion}) or a region of identical shape and area located within healthy tissue (\textsc{control}). We implemented two distinct control constructions. For the \emph{mirror} control, we reflected the tumor mask to the contralateral hemisphere. For the \emph{texture-matched} control, we translated the mask to a healthy brain location where the local gradient energy most closely matched the original tumor region. We strictly excluded any cases where the newly mapped control region extended outside the brain or overlapped with the original tumor. This exclusion criteria ensured that the perturbed lesion and control areas remained perfectly identical.

We define the specificity ratio as $\kappa = \Delta_{\textsc{lesion}} / \Delta_{\textsc{control}}$, where $\Delta$ denotes the measured distributional shift between the real and perturbed images. A ratio of $\kappa > 1$ directly indicates a stronger metric response to tumor-region perturbation. 

We utilized KID, an unbiased MMD estimator computed on InceptionV3 features using a polynomial kernel, as our primary natural-image baseline. To carefully isolate the effect of the network backbone from the choice of kernel, we also computed a kernel-matched baseline. We achieved this by applying the multi-bandwidth RBF $MMD^2$ detailed in Algorithm~\ref{alg:mmd} directly to the standard InceptionV3 features. 

\begin{figure}[pos=htbp]
\centering
\includegraphics[width=\textwidth]{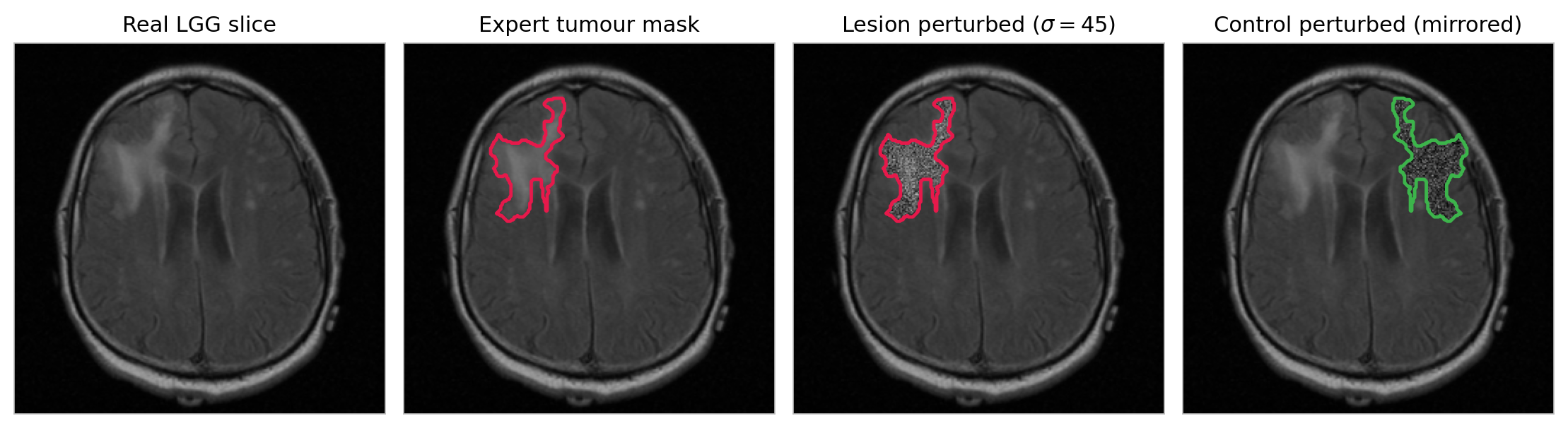}
\caption{Area-matched lesion specificity design applied to a real LGG slice. From left to right: the original image slice; the expert FLAIR abnormality mask highlighted in red; the \textsc{lesion} condition with Gaussian noise ($\sigma = 45$) injected strictly within the tumor region; and the \textsc{control} condition with identical noise injected within a mirrored, contralateral region of equal area highlighted in green. Because both regions share an identical shape and area, any observed differences in the metric response are purely attributable to the anatomical location of the perturbation.}
\label{fig:lesion_example}
\end{figure}

Each test condition comprised $N = 200$ images. We injected region-local Gaussian noise using standard deviation levels of $\sigma \in \{10, 20, 30, 45, 60, 90\}$ on the 0 to 255 intensity scale of the LGG images. We subsequently clipped the resulting perturbed intensities to the bounded $[0, 255]$ range. Figure~\ref{fig:lesion_example} illustrates both experimental conditions.

\paragraph{Results.}
When the target region was completely erased (set to zero), no evaluated metric demonstrated meaningful lesion specificity ($\kappa \approx 1.1$ to $1.3$ across all metrics). However, when we introduced region-local Gaussian noise, $M^3$-Score responded significantly more strongly to tumor-region perturbations than to control-region perturbations. In stark contrast, the Inception-based baselines responded similarly to both areas. 

\begin{figure}[pos=htbp]
\centering
\includegraphics[width=0.95\linewidth]{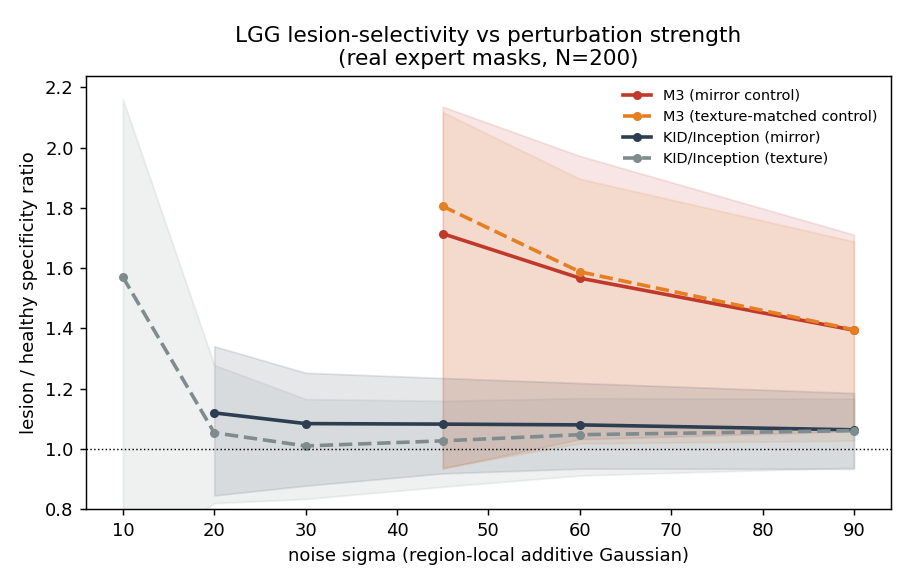}
\caption{Lesion-to-healthy specificity ratio $\kappa$ plotted as a function of the region-local noise level $\sigma$ using real LGG expert masks ($N = 200$). $M^3$-Score (red and orange lines) attains $\kappa = 1.4$ to $1.8$ for both the mirror and texture-matched controls. In contrast, the Inception-based KID (dark blue and gray lines) remains stagnant near 1.05. Shaded bands denote the bootstrap 95\% confidence intervals. Note that $M^3$-Score values falling below its lower detection threshold ($\sigma < 45$) are omitted.}
\label{fig:lgg_sweep}
\end{figure}

For noise levels $\sigma < 45$, the L12 $MMD^2$ value for $M^3$-Score dropped to zero, rendering the specificity ratio undefined. For noise levels $\sigma \ge 45$, $M^3$-Score consistently achieved a ratio of $\kappa = 1.4$ to $1.8$ for both control methodologies, with the two controls agreeing to within 0.1. Meanwhile, the Inception-based baselines stagnated between 1.03 and 1.10 across all tests (detailed in Figure~\ref{fig:lgg_sweep} and Table~\ref{tab:lgg_kernel}). 

Focusing specifically on the mirror control at the higher noise levels ($\sigma \in \{45, 60, 90\}$), the kernel-matched baseline attained $\kappa = 1.08$ to $1.10$, and standard KID attained $\kappa = 1.06$ to $1.08$. These natural-image baselines remained within 0.02 of each other at every single noise level. Conversely, $M^3$-Score achieved a much higher range of 1.39 to 1.71. This performance gap definitively confirms that the heightened sensitivity of $M^3$-Score is directly attributable to the medical-specific ViT backbone rather than the multi-bandwidth RBF kernel. Furthermore, our rigorous area-matched design and gradient-energy matching successfully eliminated confounding variables such as raw pixel magnitude and localized texture complexity.

Interestingly, the lesion specificity of $M^3$-Score gradually decreased as $\sigma$ increased, eventually approaching the low specificity range observed during complete region erasure. At $\sigma = 45$, the bootstrap confidence interval for the mirror control still included the null value of $\kappa = 1$ ($[0.94, 2.14]$). However, the lower bound successfully exceeded 1.0 at $\sigma = 60$ and 90 (reaching 1.03 and 1.06 for the mirror control, and 1.02 and 1.03 for the texture-matched control). These results demonstrate that the metric's specificity is highly tuned to subtle, clinically relevant degradations of diagnostic regions, and this targeted sensitivity is statistically significant.

\begin{table}[htbp]
\centering
\caption{Lesion-to-healthy specificity ratio $\kappa$ with bootstrap 95\% confidence intervals for the mirror control condition ($N = 200$). The kernel-matched baseline (RBF computed on InceptionV3) differs from $M^3$-Score exclusively in its backbone architecture, whereas KID differs in both the backbone and the kernel. The ``N/A'' designation indicates a noise level falling below the metric's detection threshold. Values for $\sigma = 10$ are excluded due to lack of signal.}
\label{tab:lgg_kernel}
\setlength{\tabcolsep}{4pt}
\small
\begin{tabular}{lccc}
\toprule
$\sigma$ & $M^3$-Score (RadioDINO-s16, RBF) & InceptionV3, RBF & InceptionV3, polynomial (KID) \\
\midrule
20 & N/A & $1.13 \ [0.85, 1.32]$ & $1.12 \ [0.85, 1.34]$ \\
30 & N/A & $1.09 \ [0.88, 1.26]$ & $1.08 \ [0.88, 1.25]$ \\
45 & $\mathbf{1.71} \ [0.94, 2.14]$ & $1.10 \ [0.92, 1.26]$ & $1.08 \ [0.92, 1.24]$ \\
60 & $\mathbf{1.57} \ [1.03, 1.97]$ & $1.10 \ [0.95, 1.24]$ & $1.08 \ [0.93, 1.22]$ \\
90 & $\mathbf{1.39} \ [1.06, 1.71]$ & $1.08 \ [0.96, 1.20]$ & $1.06 \ [0.93, 1.19]$ \\
\bottomrule
\end{tabular}
\end{table}

\subsection{Sensitivity to Occlusion Artifacts}
\label{sec:results_occlusion}

We examined the response of each metric to structural occlusion artifacts by placing a $64 \times 64$ zero-filled square at the center of the images. We applied this square either to the real reference images (condition B) or to the generated images (condition C), comparing each masked condition against the unmasked baseline (condition A), as summarized in Table~\ref{tab:masking}. 

Masking the real images unexpectedly reduced the $M^3$-Score fidelity distance by 39.9\%. Conversely, standard FID and CMMD dramatically increased by 211.0\% and 221.0\%, respectively. When we masked the generated images instead, FID and CMMD skyrocketed by 264.2\% and 358.7\%, whereas $M^3$-Score experienced a much more moderate increase of 63.0\%.

The massive spikes observed in FID and CMMD across both conditions indicate an overwhelming sensitivity to the artifact itself, completely independent of which dataset actually contains the distortion. In contrast, $M^3$-Score proved to be considerably more robust. Because a central square occludes both lesion and non-lesion anatomy indiscriminately, this analysis demonstrates that natural-image baselines are hypersensitive to generic image artifacts rather than meaningful clinical structures.

\begin{table}[htbp]
\centering
\caption{Impact of a central occlusion artifact ($64 \times 64$ zero-filled square) applied to the standard reference dataset ($N = 500$ across 500 subjects). Condition B masks the real images, while condition C masks the generated images. Relative changes are computed with respect to the unmasked baseline (condition A, corresponding to Tables~\ref{tab:canonical} and~\ref{tab:cmmd_fail}).}
\label{tab:masking}
\setlength{\tabcolsep}{6pt}
\small
\begin{tabular}{lccc}
\toprule
Condition & $M^3$-Score & FID & CMMD \\
\midrule
A: Real vs. generated (baseline) & $0.0732$ & $68.4$  & $0.1040$ \\
B: Real vs. masked real          & $0.0440$ & $212.7$ & $0.3339$ \\
C: Real vs. masked generated     & $0.1192$ & $249.0$ & $0.4772$ \\
\midrule
Relative change (B $-$ A) / A    & $-39.9\%$ & $+211.0\%$ & $+221.0\%$ \\
Relative change (C $-$ A) / A    & $+63.0\%$ & $+264.2\%$ & $+358.7\%$ \\
\bottomrule
\end{tabular}
\end{table}

\subsection{Sample-Size Dependence}
\label{sec:results_cv}

Two properties related to sample size were evaluated: repeatability, defined as the variation of a metric across independent draws at a fixed $N$, and comparability, defined as the stability of the mean value of a metric across different values of $N$ on the same data. For each $N$, ten independent draws were obtained. Real images were drawn by first sampling subjects and then slices, so that the reported variation reflects the construction of a new reference rather than the reshuffling of slices within a fixed set of subjects.

\paragraph{Repeatability}
FID exhibited a lower CV than M3 at every sample size below $500$ (Table~\ref{tab:cv}): $3.6\%$ compared with $6.7\%$ at $N = 100$, and $9.0\%$ compared with $13.0\%$ at $N = 25$. The unbiased MMD$^2$ estimator used by M3 and CMMD has a higher small-sample variance than the Gaussian-moment estimator of FID. At $N = 500$, the CVs of the three metrics were similar ($1.27\%$, $1.25\%$, and $1.17\%$ for M3, FID, and CMMD, respectively). Accordingly, $N \geq 100$ is recommended for M3, and $N = 500$ is preferable.

\paragraph{Comparability}
Across a twentyfold range of sample sizes, with the same real data and the same generator, the mean M3 fidelity value ranged from $0.0717$ to $0.0750$, a maximum-to-minimum ratio of $1.05$. Over the same range, the mean FID decreased from $169.5$ to $67.3$, a ratio of $2.52$. This dependence reflects the finite-sample bias of the Fr\'echet estimator, documented for natural images by \citet{chong2020effectively}. As a consequence, FID values computed with different reference sizes are not directly comparable, and the difference between them is comparable to the difference between a high-quality and a low-quality generator (Tables~\ref{tab:cv} and~\ref{tab:cmmd_fail}). The mean CMMD varied by a factor of $1.04$, indicating that the stability of M3 derives from the unbiased estimator rather than from the backbone.

These results indicate that FID offers lower variance for evaluations performed at a fixed, small sample size and compared only internally, whereas M3 is better suited to comparisons across studies, reference sizes, or published values.

\begin{table}[t]
\centering
\caption{%
  Sample-size dependence based on ten independent draws for each $N$, with real images resampled at the subject level. CV quantifies repeatability at fixed $N$ (lower is better; the lowest value in each row is shown in bold). Mean quantifies comparability across $N$.
}
\label{tab:cv}
\setlength{\tabcolsep}{4pt}
\small
\begin{tabular}{lcccccc}
\toprule
& \multicolumn{3}{c}{CV (\%)} & \multicolumn{3}{c}{Mean} \\
\cmidrule(lr){2-4}\cmidrule(lr){5-7}
$N$ & M3 & FID & CMMD & M3 & FID & CMMD \\
\midrule
$25$  & $12.99$ & $\mathbf{8.98}$ & $11.97$ & $0.0734$ & $169.5$ & $0.1019$ \\
$50$  & $9.71$  & $\mathbf{4.32}$ & $6.57$  & $0.0717$ & $132.6$ & $0.1003$ \\
$100$ & $6.73$  & $\mathbf{3.63}$ & $7.44$  & $0.0750$ & $112.0$ & $0.1038$ \\
$200$ & $5.60$  & $2.25$          & $\mathbf{2.10}$ & $0.0722$ & $88.2$ & $0.1001$ \\
$500$ & $1.27$  & $1.25$          & $\mathbf{1.17}$ & $0.0732$ & $67.3$ & $0.1028$ \\
\midrule
\multicolumn{4}{l}{Max/min ratio of the mean}
 & $\mathbf{1.05}$ & $2.52$ & $\mathbf{1.04}$ \\
\bottomrule
\end{tabular}
\end{table}

\section{Discussion}
\label{sec:discussion}
\subsection{Role of the Feature Space}
\label{sec:discussion_backbone}

The results indicate that the choice of feature space has a substantial effect on evaluation outcomes for radiology images. InceptionV3 was trained on 1,000 natural-image classes and CLIP on web-scale image-caption pairs; neither model was exposed to the joint variation of tissue intensity, pathological contrast, and acquisition characteristics that defines radiological data. On the in-domain discrimination task, RadioDINO-s16 attained a ROC-AUC of 0.819, compared with 0.555 for InceptionV3 and 0.582 for CLIP. The CMMD rank reversal (Table~\ref{tab:cmmd_fail}) provides a further example: CLIP features placed cross-modality CT closer to real brain MRI than a weaker same-modality generator, whereas RadioDINO-s16 and InceptionV3 features preserved the expected ordering.

A plausible explanation is that CLIP representations are optimized for image-text alignment and therefore emphasize high-level semantic content that is shared across modalities, rather than the intensity and texture statistics that distinguish radiological acquisitions. The present design does not separate the contribution of the training objective from those of model scale and training data, because CLIP ViT-L/14 and RadioDINO-s16 differ in all three respects; an architecture-matched comparison would be required. The reversal was observed on a single dataset and a single pair of comparison sets and should not be generalized to the use of CMMD for natural images. Moreover, the WDM-3D set differs from the FLAIR reference in MR sequence (T1-weighted) as well as in generator quality, so the reversal may partly reflect how CLIP represents a change of MR contrast. Consistent with \citet{woodland2024feature}, these findings indicate that the suitability of a backbone must be established for the target task rather than inferred from its training domain.

\subsection{Separate Evaluation Axes}
\label{sec:discussion_axes}

The DDPM results demonstrate the value of reporting fidelity and coverage separately: a low fidelity distance was accompanied by limited coverage, a combination that a single scalar cannot represent. The memorization axis provides complementary information on the reproduction of real data. The coverage experiments further show that the choice of estimator determines whether a diversity statistic responds in the expected direction to mode dropping. The $M^3$ axes complement manifold-based metrics such as $\alpha$-precision and $\beta$-recall, and $M^3$-Score is intended to be reported alongside FID. FID offers lower variance at small fixed sample sizes, whereas $M^3$ provides radiology-specific discrimination, stable values across sample sizes, and axis-specific diagnostic information.

\subsection{Limitations}
\label{sec:limitations}

\paragraph{Small-sample variance}
At sample sizes below 500, the fidelity estimate exhibited a higher CV than FID (Table~\ref{tab:cv}), and the unclipped estimator can take negative values for $N < 50$. For small evaluation sets that are compared only internally, FID provides lower variance. $M^3$ is recommended for $N \ge 100$.

\paragraph{Encoder depths}
The three depths were pre-specified according to relative position and were not optimized. The near-duplicate experiment recovered the injected fraction at all depths and therefore cannot rank them, and the direction of the mode-drop response was independent of depth. Because the threshold $\theta$ depends on the composition of the reference set, depth comparisons based on stronger near-duplicate perturbations may be confounded by reference composition. A systematic analysis of depth across tasks, with controlled reference composition, is required.

\paragraph{Scope of the permutation test}
The permutation test evaluates the null hypothesis that the real and generated distributions are identical. This hypothesis was rejected for every generated and out-of-domain set, and the test did not differentiate among them. Comparison of generators would require a null distribution derived from the variability between independent cohorts of real patients.

\paragraph{Reference composition}
The dependence of reported values on the composition of the reference set, and the relationship between the effective sample size of a slice-based reference and its number of subjects, were not quantified. These questions apply to all distributional metrics.

\paragraph{Lesion specificity}
The lesion experiment combined real expert masks with synthetic perturbations. It demonstrates that the representation responds differently to tumor tissue than to area- and texture-matched healthy tissue, but it does not establish sensitivity to clinically relevant generation errors such as hallucinated or missing lesions. Such an evaluation would require a lesion-conditioned generator assessed against the same masks.

\paragraph{Datasets and baselines}
Validation was performed primarily on brain MRI from a single public dataset at $N = 500$. The in-domain DDPM was trained on all BraTS subjects, so its evaluation against the reference set does not measure generalization to unseen subjects, and the cross-modality and far-domain sets were synthetic rather than real images. Validation on larger, multi-site datasets and on real CT with ground-truth pathology annotations remains to be performed, and a quantitative comparison with FRD was not included. $M^3$ values for non-radiology images should not be interpreted as general measures of visual distance (Section~\ref{sec:results_ood}).

\section{Conclusion}
\label{sec:conclusion}

This study introduces the $M^3$-Score, a robust evaluation framework designed specifically for generative radiology image models. By decoupling performance into three independent axes (fidelity, memorization, and coverage) computed from pre-specified blocks of a frozen RadioDINO-s16 encoder, and by enforcing rigorous subject-level reference set construction, $M^3$-Score successfully overcomes the critical failure modes of natural-image metrics. 

Empirically, our findings demonstrate that $M^3$-Score correctly ranks datasets by degradation severity on BraTS brain MRI, yields zero shift on subject-disjoint real data, and uncovers hidden model limitations—such as an unconditional DDPM combining a low fidelity distance with a restricted coverage of only $38\%$. Furthermore, our real-radius coverage estimator behaves monotonically under mode dropping across all transformer blocks, avoiding the paradoxical behaviors observed with traditional $k$-NN recall. Compared to general-purpose representations like InceptionV3 and CLIP, RadioDINO-s16 provides vastly superior discrimination between real and generated medical scans while avoiding the rank-reversal vulnerabilities exhibited by CMMD. Additionally, $M^3$-Score demonstrates exceptional stability across wide sample-size fluctuations and exhibits heightened sensitivity to clinically vital pathological structures, outperforming Inception-based baselines in isolating tumor-region perturbations from healthy tissue. 

Ultimately, $M^3$-Score bridges the gap between technical generative evaluation and clinical utility. Future work will focus on the systematic optimization of encoder depths, establishing empirical null distributions derived from independent real-patient cohorts, investigating the nuanced effects of reference composition, validating the framework on large-scale multi-site CT datasets, and extending the evaluation protocol to fully lesion-conditioned generative architectures.

\section*{Declaration of competing interest}
The authors declare that they have no known competing financial interests or personal relationships that could have appeared to influence the work reported in this paper.

\section*{Declaration of generative AI and AI-assisted technologies in the manuscript preparation process}
During the preparation of this work, the authors used AI-assisted technologies to improve sentence structure and clarity. After using this tool, the authors reviewed and edited the content as needed and take full responsibility for the content of the published article.

\section*{Data availability}
The primary datasets used in this study are publicly available: BraTS 2021 \citep{menze2014multimodal, bakas2018identifying, baid2021rsna}, the LGG MRI dataset \citep{buda2019association}, and the Kaggle Retinal Fundus Images dataset. Pretrained weights for RadioDINO-s16 \citep{zedda2025radiodino} are publicly available on Hugging Face (\url{https://huggingface.co/Snarcy/RadioDino-s16}). Complete source code, configuration files, evaluation scripts, and pipelines required to reproduce every table and figure in this study will be released publicly upon publication.

\section*{CRediT authorship contribution statement}
\textbf{Sathiyamohan Nishankar:} Conceptualization, Methodology, Software, Formal analysis, Investigation, Visualization, Writing -- original draft.
\textbf{Pubudu Sanjeewani:} Validation, Writing -- review \& editing.
\textbf{Asanka Perera:} Supervision, Validation, Writing -- review \& editing.

\bibliographystyle{model1-num-names}

\bibliography{refs}

\end{document}